\documentclass{article}

\PassOptionsToPackage{sort&compress}{natbib}

\usepackage[final]{corl_2026}       %

\usepackage{amsmath}
\usepackage{amssymb}
\usepackage{amsfonts}
\usepackage{amsbsy}

\usepackage{xspace}
\usepackage{url}
\usepackage{subcaption}
\usepackage{tabularx}
\usepackage{graphicx}
\usepackage{float} %
\usepackage{placeins} %
\usepackage{multirow}
\usepackage{xparse}
\usepackage{nicefrac}
\usepackage{microtype}
\usepackage{enumerate}
\usepackage{setspace}

\usepackage{enumitem}
\usepackage{wrapfig}
\usepackage{needspace} %
\usepackage{booktabs}
\usepackage{tikz}
\usepackage{array}
\usepackage{caption}
\usepackage{algorithm}
\usepackage{algpseudocode}
\usepackage{makecell}
\usepackage[table]{xcolor}
\usepackage{enumitem}

\graphicspath{{../}{./}}

\newcommand{\methodname}{\textsc{maestro}\xspace}
\newcommand{\methodfullname}{\textbf{M}anagerial \textbf{A}gent for \textbf{E}xecuting \textbf{S}ensorimotor \textbf{T}asks in \textbf{RO}botics\xspace}

\newcommand{\hform}[1]{\textsf{\textbf{#1}}}

\makeatletter
\newcommand{\fixedsubsection}{%
  \@startsection{subsection}{2}{\z@}%
    {-1.8ex \@plus -0.5ex \@minus -0.2ex}%
    {0.8ex}%
    {\normalsize\bf\raggedright}%
}
\makeatother

\title{\methodname: Orchestrating Robotics Modules with Coding Agents for Zero-Shot Generalist Robots}

\author{
  Junyao Shi$^{*}$, Rujia Yang$^{*}$, Kaitian Chao$^{*}$, Selina Bingqing Wan \\
  \textbf{Yifei Shao, Jiahui Lei, Jianing Qian, Long Le} \\
  \textbf{Pratik Chaudhari, Kostas Daniilidis, Chuan Wen, Dinesh Jayaraman} \\
  \normalfont $^{*}$\,equal contribution \\
  \normalfont University of Pennsylvania \\
  \normalfont \href{https://maestro-robot.github.io/}{maestro-robot.github.io}
}
\hypersetup{
  pdfauthor={Junyao Shi, Rujia Yang, Kaitian Chao, Selina Bingqing Wan, Yifei Shao, Jiahui Lei, Jianing Qian, Long Le, Pratik Chaudhari, Kostas Daniilidis, Chuan Wen, Dinesh Jayaraman},
  pdftitle={MAESTRO: Orchestrating Robotics Modules with Coding Agents for Zero-Shot Generalist Robots}
}

\begin{document}

\maketitle

\begin{abstract}
\frenchspacing %
\space
Today's dominant route to generalist robots scales robotics data to train end-to-end policies.
We instead use a vision-language model (VLM) as a coding agent that writes robot programs using a curated library of perception, planning, and control tools.
\methodname{}'s orchestration harness packages multi-step perception and action procedures into callable tools and guides their use through a plan--act--replan loop, using visual and execution feedback to verify outcomes and recover from failures.
With this tool library and harness, \methodname{} matches or surpasses state-of-the-art vision-language-action (VLA) models zero-shot in our tabletop comparison and performs diverse mobile manipulation tasks, without collecting new teleoperation data or invoking a VLA tool. Ablations show that expanding the tool library improves average task progress with the full harness, while adding tools without it does not reliably improve progress.
\space
Case studies illustrate \methodname{}'s potential for autonomous code improvement across trials, faster execution with VLA tools, and automatic robot data generation.%
\space
We show that expanding robotics tool libraries together with their orchestration harnesses offers a complementary route to general-purpose robots alongside scaling end-to-end robot training data.
\space
Project site: \href{https://maestro-robot.github.io/}{maestro-robot.github.io}. %
Code: \href{https://github.com/penn-pal-lab/maestro}{github.com/penn-pal-lab/maestro}.
\end{abstract}


\section{Introduction}
\suppressfloats[t] %

The prevailing view in robotics holds that general-purpose capabilities require training a single end-to-end model on massive robotics-specific datasets gathered through teleoperation~\cite{black2024pi_0,team2025gemini,bjorck2025gr00t,geminirobotics15}. The scarcity of such data, relative to the text and images available to language and vision models, is widely cited as why robotic systems lag behind, motivating large efforts to scale data collection for vision-language-action (VLA) models.
\begin{figure}[t]
  \centering
  \includegraphics[width=0.9\textwidth]{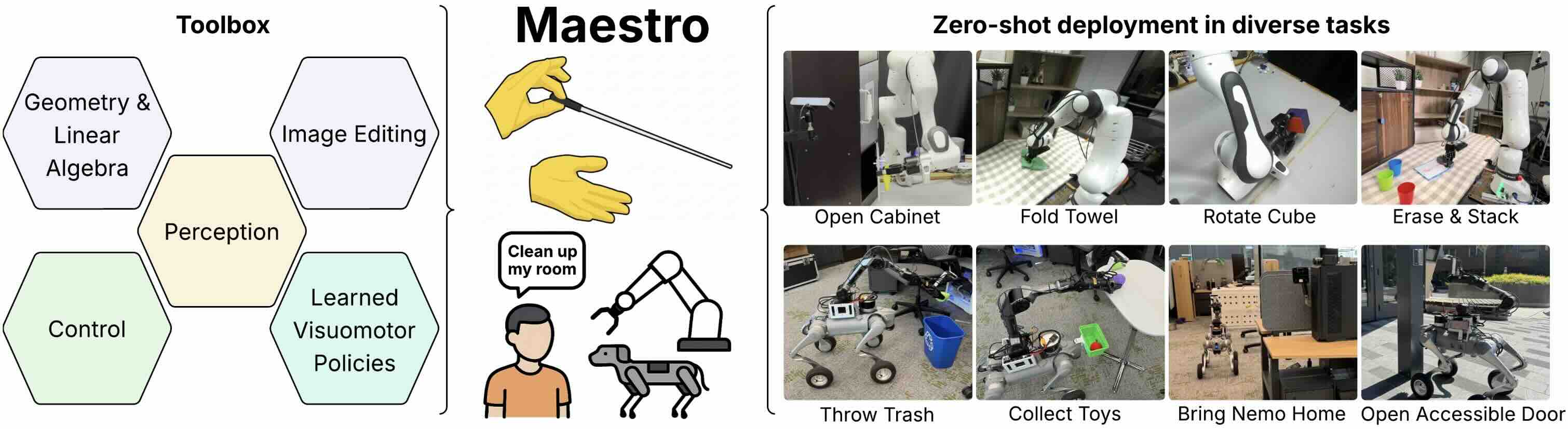}
  \caption{\textbf{\methodname} receives language instruction and leverages a set of tools to complete diverse tasks in a zero-shot setting.}
  \label{fig:concept}
  \vspace{-1mm}
\end{figure}

We take the view that robotics data to inform a generalist robot need not begin and end with such ``\textit{observations-in, actions-out}'' data.
\space
In particular, much useful robotics knowledge is readily available through a diverse collection of tools to perform robotics-relevant functions such as object pose estimation, SLAM, semantic object category recognition, or path planning. Each such module represents the cumulative engineering effort of the robotics research community, often over several decades, and informed by various data islands specific to their respective functions.

We introduce \methodname{}, a framework for tabletop and mobile manipulation that assembles these islands into one working \textit{generalist robotics} system. A vision-language model (VLM) acts as a coding agent, orchestrating a curated tool library through a closed-loop harness that monitors and recomposes tools. By replacing the task-specific bespoke human labor of coding and orchestrating specific sequences of functions with dynamically generated code on a VLM, this \textit{agentic} approach scales modular ``old-fashioned''~\cite{goldberg2025gofe} approaches to the generalist robotics setting.

Without collecting new teleoperation data, \methodname{} matches or surpasses the evaluated VLA baselines in our zero-shot tabletop comparison and performs diverse mobile manipulation tasks. Our evaluation spans pick-and-place, deformable- and articulated-object manipulation, spatial reasoning, tool use, object-affordance reasoning, semantic and memory challenges, long-horizon manipulation and loco-manipulation, and active exploration. In our ablations, the full toolset and harness achieve 76.9\% average task progress. Removing the harness reduces this to 33.0\% (43.9 percentage points lower); retaining the full harness but removing the added robotics tools reduces it to 32.3\% (44.6 points lower; Fig.~\ref{fig:scaling}).

\methodname{} also inherits the advantages of manually engineered modular systems: interpretability, debuggability, and extensibility. New tools, including VLAs, integrate with minimal effort, and adapting to new embodiments or improving from limited experience requires only localized code edits, not retraining or large-scale data collection.

These advantages currently come at higher latency and computational cost than end-to-end VLA policies, a limitation that will ease as the VLM inference stack improves.
We also show another solution route: \methodname{} can generate robot data automatically (App.~\ref{app:tether}), which could support distillation into lightweight imitation policies~\cite{goldberg2025gofe}.%
\space
Large VLMs, including GPT-6 Astra~\cite{galbot2026astra,zhang2026unexpected}, are now being used to generate robot actions directly. They offer broad reasoning capabilities but have high inference latency and token usage. Smaller VLA policies provide faster control but can struggle with open-ended, reasoning-intensive tasks.
 \space
\methodname{} explores an intermediate approach: retain VLM reasoning for planning and program synthesis, while using existing or pretrained skills for perception and control.

In summary, our contributions are:

\begin{enumerate}[leftmargin=*,topsep=3pt,itemsep=2pt,parsep=0pt,partopsep=0pt]
    \item \hspace{0pt}We introduce \methodname{}, a framework that makes a VLM coding agent into a generalist robot policy by combining a curated robotics tool library with an orchestration harness for composing tools, verifying outcomes, and recovering from execution failures.
    \item
\space
We show that \methodname{} matches or surpasses the evaluated VLA baselines and our code-as-policies baseline zero-shot in our tabletop comparison, without new teleoperation data or VLA tool use.%
 \space
We also demonstrate \methodname's capabilities in long-horizon mobile manipulation.
    \item Our systematic ablations show that expanding the tool library improves average task progress with the full harness, while adding tools without it does not reliably improve progress (Fig.~\ref{fig:scaling}).
    \item Our case studies demonstrate \methodname{}'s potential to autonomously improve its code across trials, execute tasks faster with VLA tools, and generate robot data.
\end{enumerate}

Rollout videos for all real-world experiments and an interactive agent dashboard (perception, code, and rollout views) are on our \href{https://maestro-robot.github.io/}{project site}.

\section{Related Work}

Modern vision-language-action (VLA) foundation models for robotics typically start from pre-trained VLM/LLM backbones and fine-tune on large quantities of robotics-specific data~\cite{Zawalski24-ecot,chen2025ecot-lite,shi2025hirobot,zhao2025cotvla,team2025gemini,pi2025pi05,gr00t_n1_5,lee2025molmoactact,geminirobotics15}, including hierarchical and reasoning-augmented variants~\cite{shi2025hirobot,chen2025ecot-lite}; in all of these, action selection still flows through a fine-tuned VLA. We are instead interested in directly using off-the-shelf VLMs and LLMs in the robotic control loop.
Among such applications of ``VLMs and LLMs as policies'', one approach generates low-level robotic actions directly~\cite{Kwon2024zeroshot,galbot2026astra,zhang2026unexpected}, while another generates ``code as policies''.
CaP~\cite{codeaspolicies2022} first demonstrated a text-based LLM orchestrating a small set of perception and control APIs into a program, but the generated program is static and open-loop~\cite{codeaspolicies2022,singh2023progprompt,ha2023scalingup}. More recent work closes the loop with VLMs, generating and revising code from visual feedback within a trial~\cite{duan2024manipulateanything,zhi2025comerobot,shi2025hirobot}.

\methodname{} studies how a VLM-based harness can orchestrate a broad robotics toolset across manipulation tasks, with ablations evaluating the roles of the toolset and harness. This modular approach is similar to the good old-fashioned engineering (GOFE) discussion~\cite{goldberg2025gofe}, which emphasizes combining model-based robotics with learned components.

Direct VLM-based robot control remains challenging, particularly for precise and dexterous manipulation. Many leading VLA systems achieve broad manipulation capabilities but rely on substantial, costly real-world teleoperation~\cite{black2024pi_0,pi2025pi05,team2025gemini,bjorck2025gr00t,lee2025molmoactact,geminirobotics15}; complementary data sources include simulation~\cite{dalal2024manipgen,lum2024dextrahg,lin2025recipe} and human videos~\cite{shi2025zeromimic,yang2025egovla,zhu2025emma}. The original zero-shot code-generation setup in Gemini Robotics~\cite{team2025gemini,geminirobotics15} illustrates these challenges: its reported zero-shot dress-folding trials were unsuccessful, with insufficiently precise grasp generation identified as the main difficulty. We adopt the Gemini Robotics agent as our \emph{Base CaP} reference system and the state-of-the-art $\pi_{0.5}$ model~\cite{pi2025pi05} as a strong VLA baseline; Sec.~\ref{sec:background} describes how Base CaP works and how \methodname{} departs from it by composing external robotics models as callable tools rather than relying on the VLM for perception. We show that scaling the \emph{right set of tools and harnesses} for a coding agent is a complementary axis to scaling robot data, and \methodname{} can further call VLAs as tools where they excel. App.~\ref{app:extended_related_work} extends this discussion to VLMs and LLMs as modular components and to concurrent code-as-policies efforts.

Concurrent code-as-policies efforts include CaP-X~\cite{fu2026capx}, which benchmarks and improves embodied coding agents, and GaP~\cite{chen2026gap}, which builds graph policies from modular robot skills. \methodname{} focuses on a broad robotics toolset and its orchestration harness for real-world generalist manipulation, with ablations that examine their respective roles.

\section{Method}

We call our approach \methodfullname, or \textbf{\methodname} for short.
\methodname is a robotic system centered on an agent that writes and executes code to leverage a toolkit spanning perception, geometry, control, pre-trained policies, and image editing.

\begin{figure*}[!htbp] %
  \centering
  \includegraphics[width=0.7\textwidth]{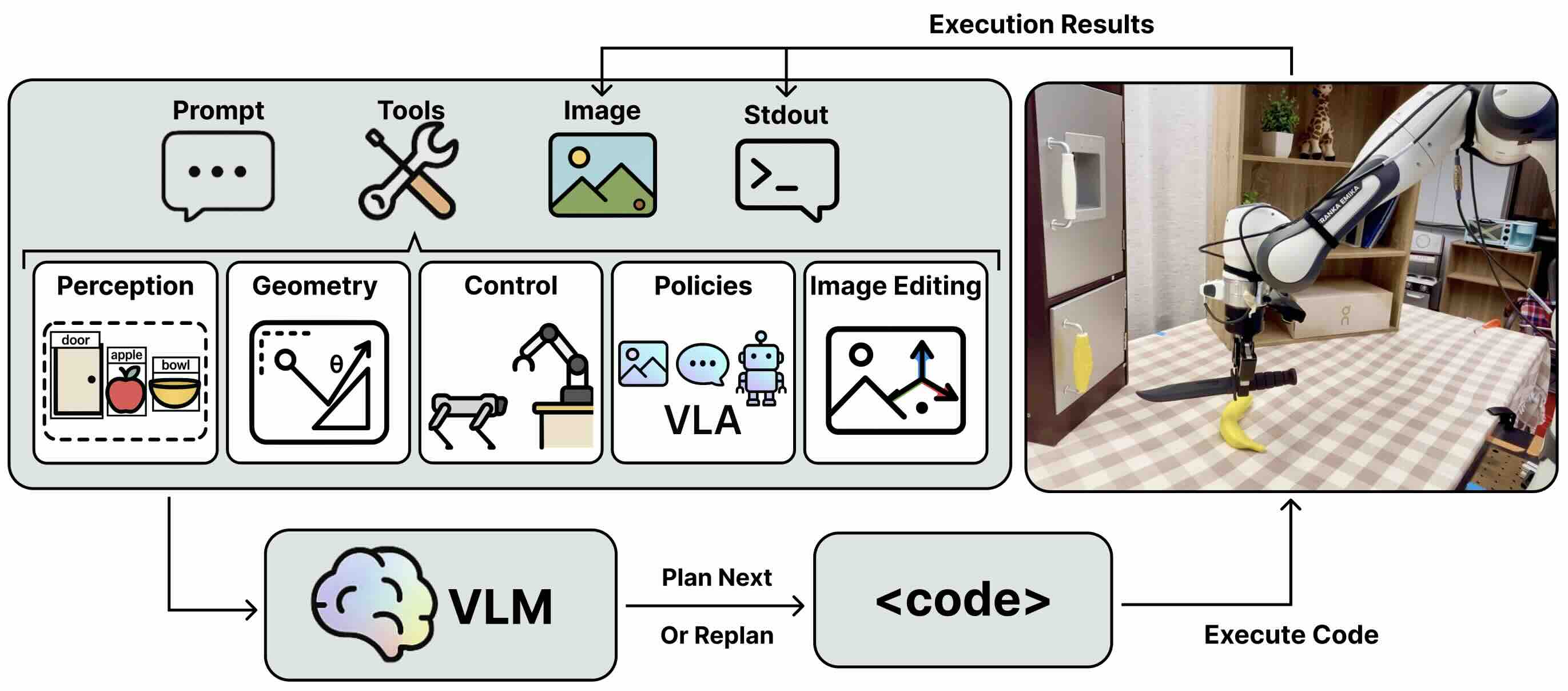}
  \caption{Given prompt and images, VLM plans by writing and executing code that integrates perception, spatial reasoning, control, learned visuomotor policies, and image editing. Execution results (images and stdout) provide feedback for reacting and replanning, forming a closed-loop perception–action–learning cycle. This enables adaptive long-horizon manipulation, as shown at right (instruction: Grasp knife by the handle and cut banana in the middle).}
  \label{fig:method}
\end{figure*}

To specify a task,
\methodname receives a system prompt, a scene image, and task instructions.
This design forms an adaptive perception–action–learning loop; we begin from the Base-CaP starting point (Sec.~\ref{sec:background}) and build up from there. An overview of this loop and the API is shown in Fig.~\ref{fig:method}, with the per-trial execution loop formalized in Algorithm~\ref{alg:maestro} and the full execution-and-evolution loop in Algorithm~\ref{alg:maestro_evolution} (both in App.~\ref{app:method_details}).

\subsection{Background: Base CaP coding agents}
\label{sec:background}

\methodname{} builds on the design pattern we call \emph{Base CaP},
our name for a representative closed-loop, code-as-policies coding agent for manipulation, exemplified by the Gemini Robotics agent~\cite{team2025gemini, geminirobotics15}.%
 \space
A Base-CaP system gives a VLM a small inventory of robot APIs and has it write and execute Python in a plan-react-replan loop: decompose the instruction, emit code for a substep, observe the resulting image and \texttt{stdout}, then advance or rewrite. Crucially, it relies on the VLM's own \emph{default} abilities: a three-tool inventory in which the VLM itself produces bounding-box detections, 2D points, and top-down grasp poses, alongside basic Cartesian and gripper control~\cite[p.~11, Fig.~12]{team2025gemini}. This places a hard ceiling on the system, which perceives and grasps only as well as a generalist VLM can. We take this configuration as our starting point, the S0 baseline of the scaling study (Sec.~\ref{sec:scaling_ablation}) and the ``Shared CaP pattern'' column of Table~\ref{tab:comparison}. The rest of this section keeps this closed-loop structure but scales it with curated best-quality modules (Sec.~\ref{sec:modules}) and the robotics-specific orchestration harness they require (Sec.~\ref{sec:monitor}); concurrent code-as-policies efforts that converge on related ideas are discussed in App.~\ref{app:extended_related_work}.

\subsection{Scaling the tool library: from Base CaP (S0) to \methodname{} (S4)}
\space
\label{sec:modules}

For tabletop manipulation, we experiment with the widely used DROID~\cite{khazatsky2024droid} platform visualized in Fig.~\ref{fig:concept}, a 7-DoF Franka Panda robotic arm equipped with a Robotiq 2F gripper, supported by a wrist-mounted camera and a third-person camera. For this setup,
\methodname{}'s tool library is a cumulative progression over the Base-CaP starting point that mirrors the scaling study (Sec.~\ref{sec:scaling_ablation}): each stage replaces a capability the Base-CaP VLM handles with its default abilities (\textbf{S0}: 2D pointing and bounding boxes, imagined top-down grasps, Cartesian and gripper control) by the curated, best-quality module for that capability. \textbf{S1} adds a coarse-to-fine perception hierarchy (Grounded-SAM segmentation, FoundationStereo~\cite{wen2025foundationstereo} metric depth, and ReKep-inspired task-relevant keypoints~\cite{huang2024rekep}); \textbf{S2} adds GraspGen~\cite{murali2025graspgen} 6-DoF grasping and cuRobo~\cite{curobo_report23} collision-free motion planning; \textbf{S3} adds geometry primitives (vector and rotation operations) that supply the metric spatial reasoning current VLMs lack; and \textbf{S4} treats the $\pi_{0.5}$~\cite{pi2025pi05} VLA as a callable policy for short in-distribution skills, run under the monitoring harness of Sec.~\ref{sec:monitor}. For the mobile manipulator the same library is extended with embodiment-specific modules (base state estimation, active perception, navigation, a fine-grained ``nudge'' tool, and a semantic map). Table~\ref{tab:comparison} contrasts the full library with Base CaP, and App.~\ref{app:toollib_stages} (with Sec.~\ref{sec:franka_modules} and Sec.~\ref{sec:mobile_modules}) gives the per-stage and per-module details.

\fixedsubsection{Robotics-specific orchestration: the harness}
\label{sec:monitor}

A robotics agent must handle unreliable tool results: perception is noisy, grasps slip, motions are blocked, and learned policies stall.%
 \space
The orchestration problem is thus not just selecting tools but recovering when \emph{any} tool fails, by re-perceiving, re-planning, or composing a different tool. We use \emph{harness} for the interface layer that encodes how to use the raw robotics modules correctly, knowledge the VLM cannot reconstruct on its own.
The harness must account for an open physical world the agent does not control, and provide ways to visualize and verify structured outputs such as 6-DoF grasp poses and point clouds before acting on them.%
 \space
Designing this harness is central to \methodname{}, and removing it collapses the without-harness curve of Sec.~\ref{sec:scaling_ablation} even as tools are added.

Concretely, we organize \methodname{}'s tool interfaces and execution support into five recurring design principles; Table~\ref{tab:comparison} lists concrete additions over our reproduced Base CaP baseline:
 \space
(\hform{a})~\textbf{encapsulation}, bundling an error-prone upstream-to-downstream module chain behind a single call; (\hform{b})~\textbf{procedural knowledge}, injected into the prompt as priorities, guardrails, and worked composition examples; (\hform{c})~\textbf{closed-loop control}, monitoring, success detection, and replanning;
(\hform{d})~\textbf{infrastructure}, coordinating shared resources and starting and stopping policy execution (App.~\ref{app:harness_details});
 \space
and (\hform{e})~\textbf{feedback}, rendering otherwise-unreadable tool outputs (e.g., predicted points and 6-DoF poses) back into the VLM's observation so it can verify them. We describe the harness each tool cluster (S1--S4) requires.

\textbf{Per-cluster harness (S1--S4).} \emph{S1 perception:} localizer-priority guidance and a detect-once-and-cache rule that avoids repeating an ${\sim}30$\,s scan (\hform{b},~\hform{d}), active perception for informative views (\hform{d}), image-editing feedback that renders predicted points and 6-DoF poses back into the observation (\hform{e}), and an encapsulated keypoint tool behind a single call (\hform{a}).
\emph{S2 grasp and control:} encapsulation (\hform{a}) through a \texttt{get\_grasp\_pose}/\texttt{plan\_grasp} pair that combines
 \space
active perception, segmentation, depth, GraspGen, gripper-flange calibration, and collision-free execution,
with a sanity-check guardrail (\hform{b}).%
 \space
\emph{S3 geometry:} purely procedural (\hform{b}), in-context worked examples of composing vectors and rotations. \emph{S4 VLA:} \texttt{execute\_with\_pi05} runs $\pi_{0.5}$ under a parallel ${\sim}2$\,Hz Qwen-2.5-VL monitor that interrupts on success or replan (\hform{a},~\hform{c}),
with a resource manager coordinating exclusive camera access and starting and stopping policy execution (\hform{d})
; this monitor is precisely what turns a non-terminating policy into a tool the agent can call and reliably stop. App.~\ref{app:method_details} details each.

Underlying all of these is the closed-loop control loop (\hform{c}): the plan-react-replan cycle through which every recovery happens. This loop is shared with Base CaP and supports \methodname{}\textquotesingle s per-cluster harness, and we describe it next.

Given a task instruction and an image at the start of an episode, \methodname{} first \textbf{\textit{plans}}, decomposing the task into substeps and generating code for the first; it then \textbf{\textit{reacts}}, using the instruction, code output, robot state, and post-execution images to judge whether the subgoal was achieved; if so it \textbf{\textit{plans}} the next substep, and if not it \textbf{\textit{replans}}, diagnosing the failure and rewriting code for the same substep. This \textbf{\textit{plan}--\textit{react}--\textit{replan}} loop lets \methodname{} continuously adapt to the environment and its own mistakes until the task completes (Alg.~\ref{alg:maestro}, Fig.~\ref{fig:prompt}, App.~\ref{app:method_details}; interactive replay on our \href{https://maestro-robot.github.io/}{project site}).

Fig.~\ref{fig:replan} (App.~\ref{app:method_details}) traces one such iteration on the whiteboard-erase task: the react step (Alg.~\ref{alg:maestro}) localizes a failed eraser grasp and rewrites one line of \texttt{actor\_code} to lower the grasp height before the retry succeeds. The loop closes at the code level rather than the retry level: a post-execution image and the robot state let the VLM localize the failure and apply a code edit instead of re-executing.

\noindent\textbf{Monitor reliability.} A separate human audit of 372 non-idle step-level success/failure judgments across 61 logged trials found 89.8\% accuracy, 85.0\% precision, and 99.0\% recall for success detection. App.~\ref{app:monitor_audit} details the audit.

\textbf{Cross-trial evolution.} \methodname{} can also autonomously improve across repeated trials on the same task: after each trial it feeds a structured execution record (generated code, reasoning trace, and a post-execution root-cause analysis) back to the VLM as in-context history, producing targeted corrections that compound without human intervention\label{sec:evolution} (formalized in Alg.~\ref{alg:maestro_evolution}; open-cabinet case study in App.~\ref{app:evolution}, results in App.~\ref{sec:evolution_results}).

\section{Experiments}

Our experiments evaluate \methodname{}'s zero-shot manipulation capabilities and the contributions of its tools and harness. We describe the real-world setup (Sec.~\ref{sec:experiment_setup}), compare against VLA and Base CaP baselines on tabletop tasks (Sec.~\ref{sec:tabletop_results}), and evaluate mobile manipulation (Sec.~\ref{sec:mobile_results}).
We then analyze the tools and harness through ablations (Sec.~\ref{sec:scaling_ablation}) and present case studies of cross-trial code improvement, VLA tool use, and autonomous data generation (Sec.~\ref{sec:beyond_single_trial}).%
 \space
Additional analyses examine individual module contributions, tool use, failure modes, and monitor reliability (Apps.~\ref{app:loo_ablation}--\ref{app:failure_cases} and~\ref{app:monitor_audit}).

\begin{figure}[!t]
  \setlength{\abovecaptionskip}{2pt}
  \setlength{\belowcaptionskip}{0pt}
  \centering
  \setlength{\tabcolsep}{1pt}
  \renewcommand{\arraystretch}{0.8}
  \begin{tabular}{*{10}{c}}
    \multicolumn{5}{l}{\parbox[t]{0.47\linewidth}{\footnotesize\raggedright\textbf{Pick–Place:} pick up item and put into the bowl}} &
    \multicolumn{5}{l}{\parbox[t]{0.47\linewidth}{\footnotesize\raggedright\textbf{Deformable object:} fold cloth corners into center}} \\
    \includegraphics[width=0.095\linewidth]{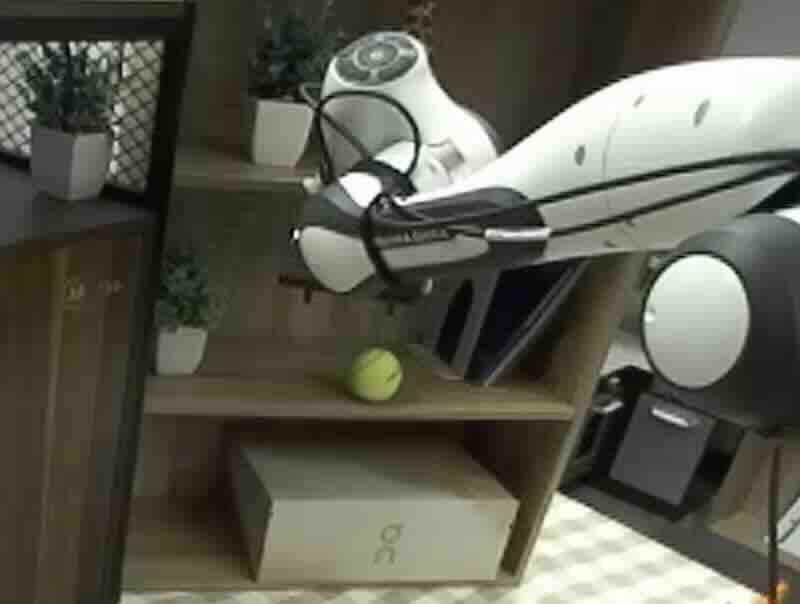} &
    \includegraphics[width=0.095\linewidth]{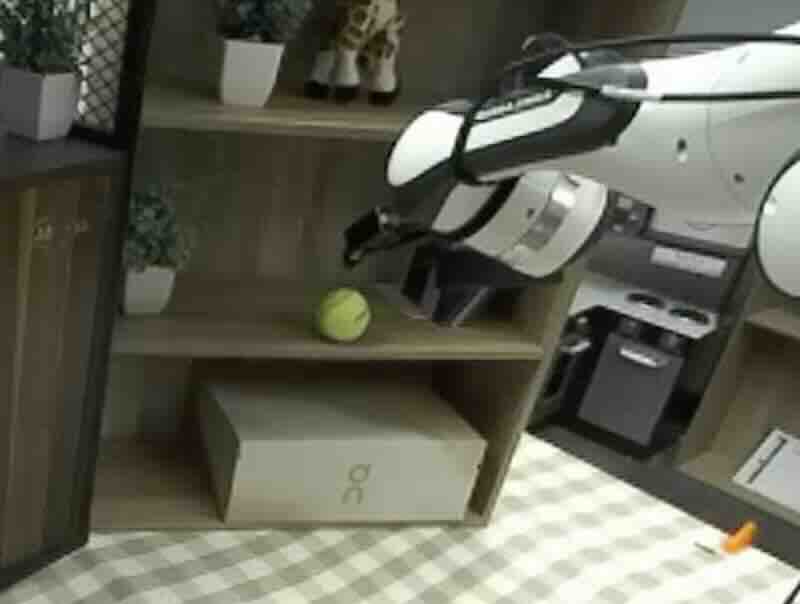} &
    \includegraphics[width=0.095\linewidth]{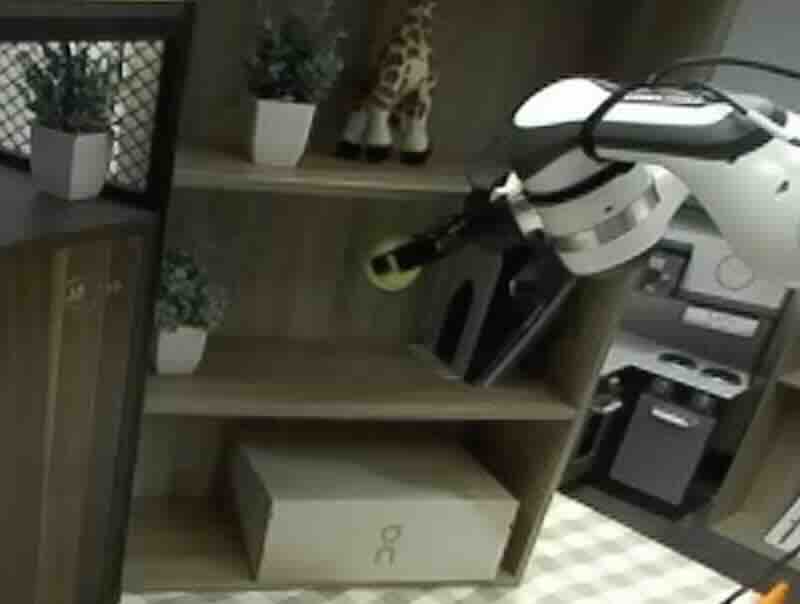} &
    \includegraphics[width=0.095\linewidth]{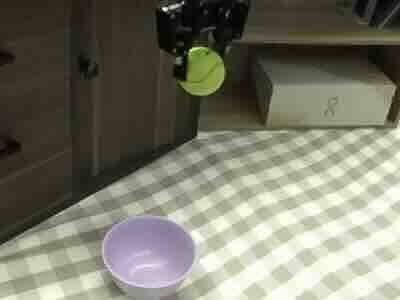} &
    \includegraphics[width=0.095\linewidth]{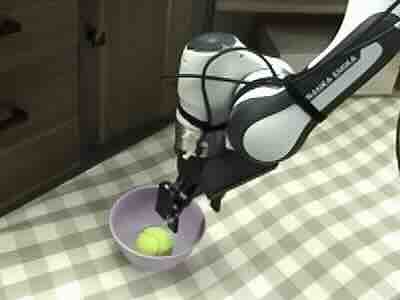} &
    \includegraphics[width=0.095\linewidth]{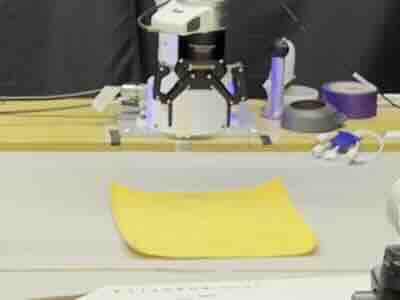} &
    \includegraphics[width=0.095\linewidth]{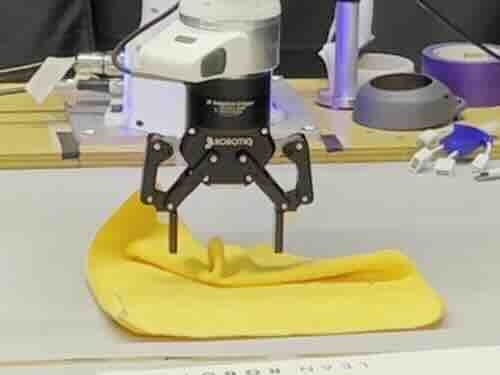} &
    \includegraphics[width=0.095\linewidth]{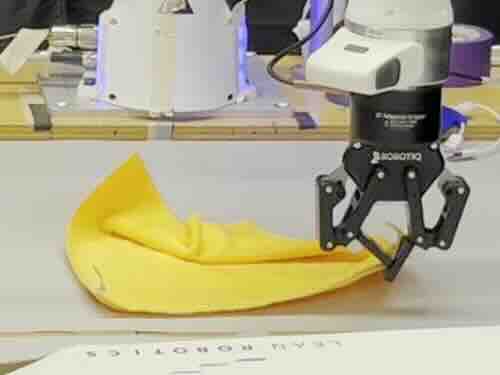} &
    \includegraphics[width=0.095\linewidth]{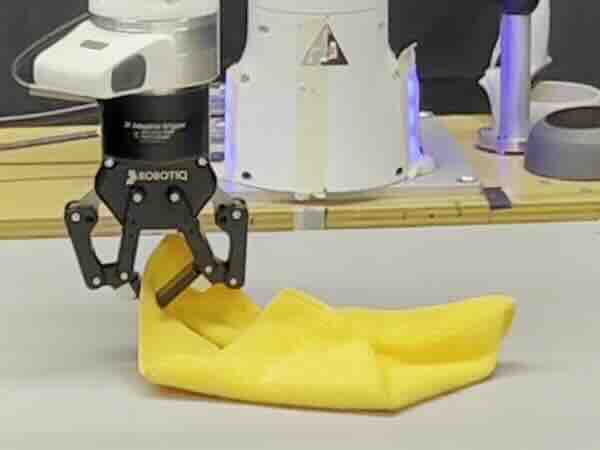} &
    \includegraphics[width=0.095\linewidth]{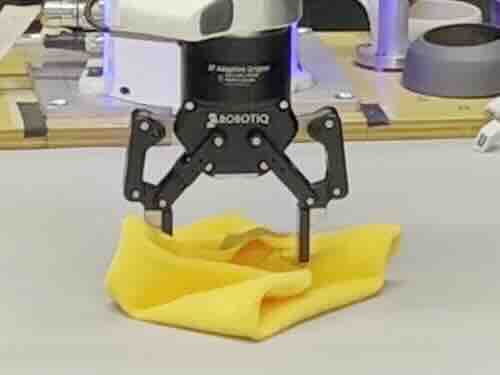} \\
    \multicolumn{5}{l}{\parbox[t]{0.47\linewidth}{\footnotesize\raggedright\textbf{Articulated object:} open cabinet}} &
    \multicolumn{5}{l}{\parbox[t]{0.47\linewidth}{\footnotesize\raggedright\textbf{Spatial reasoning:} rotate cube purple side up}} \\
    \includegraphics[width=0.095\linewidth]{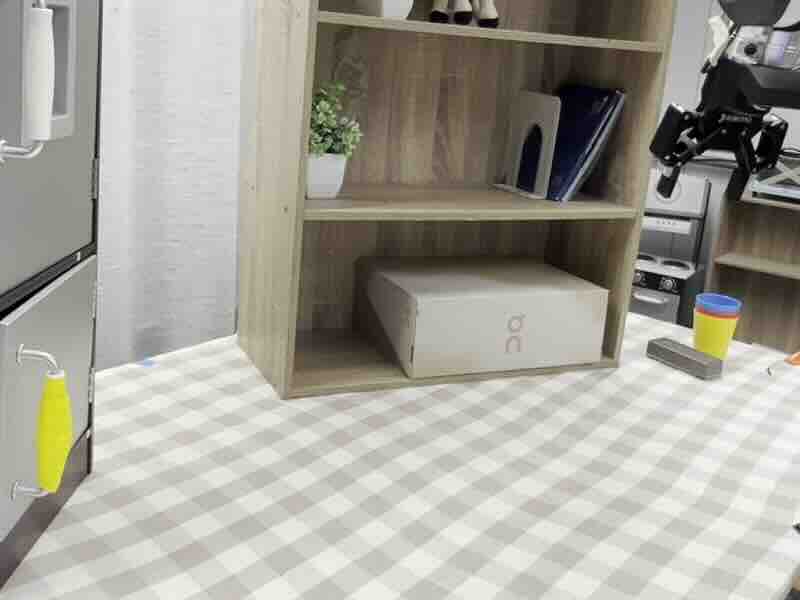} &
    \includegraphics[width=0.095\linewidth]{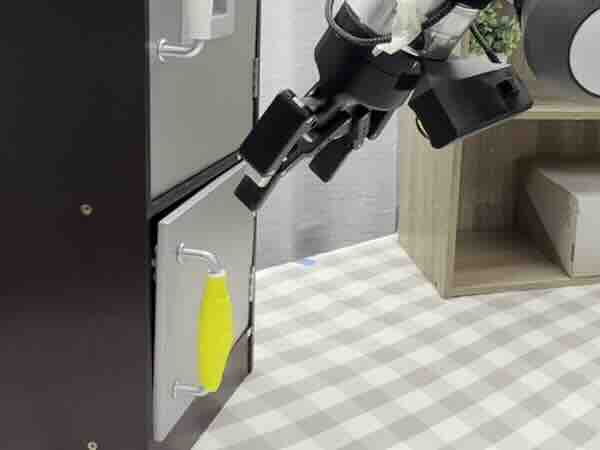} &
    \includegraphics[width=0.095\linewidth]{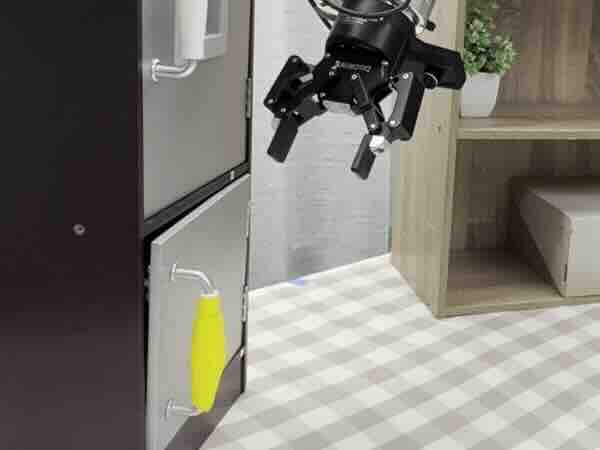} &
    \includegraphics[width=0.095\linewidth]{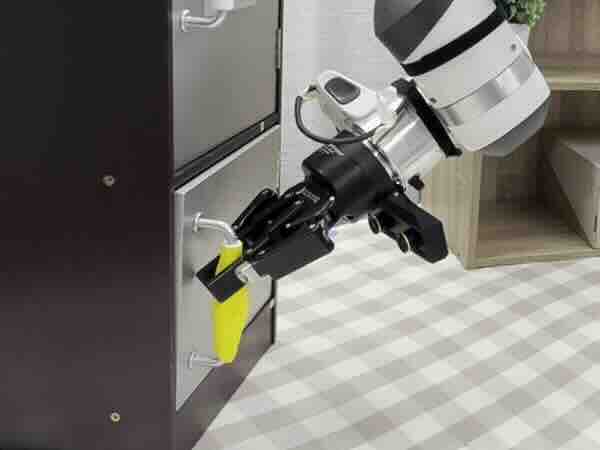} &
    \includegraphics[width=0.095\linewidth]{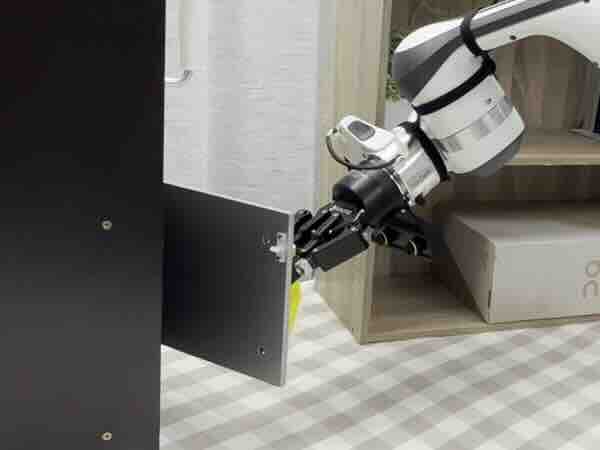} &
    \includegraphics[width=0.095\linewidth]{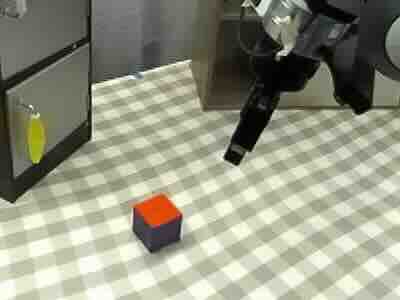} &
    \includegraphics[width=0.095\linewidth]{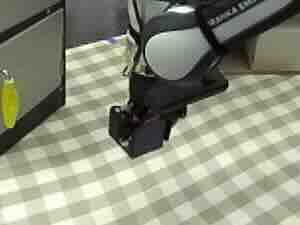} &
    \includegraphics[width=0.095\linewidth]{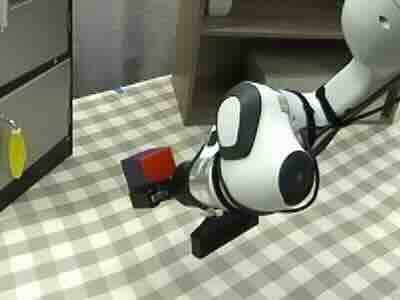} &
    \includegraphics[width=0.095\linewidth]{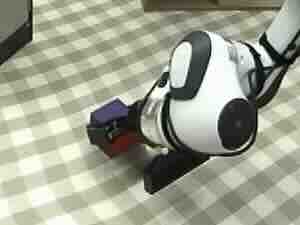} &
    \includegraphics[width=0.095\linewidth]{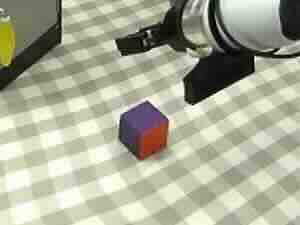} \\
    \multicolumn{5}{l}{\parbox[t]{0.47\linewidth}{\footnotesize\raggedright\textbf{Tool use:} cut banana with knife}} &
    \multicolumn{5}{l}{\parbox[t]{0.47\linewidth}{\footnotesize\raggedright\textbf{Object affordance:} hang mug on mug holder}} \\
    \includegraphics[width=0.095\linewidth]{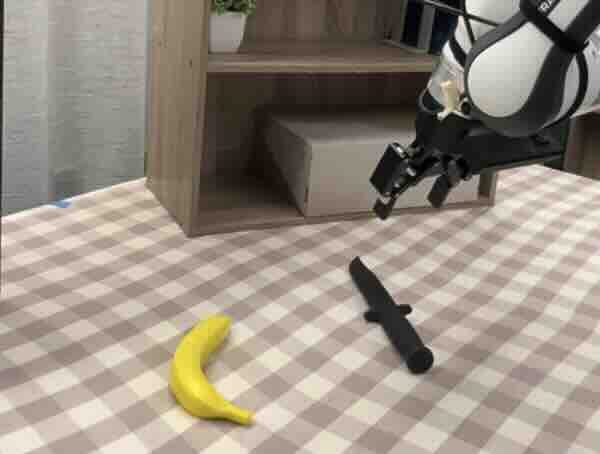} &
    \includegraphics[width=0.095\linewidth]{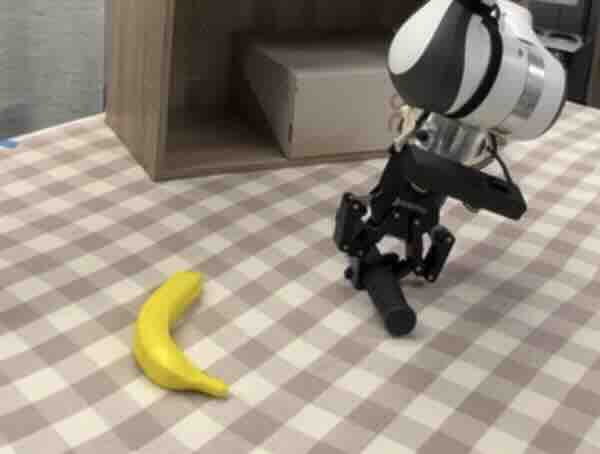} &
    \includegraphics[width=0.095\linewidth]{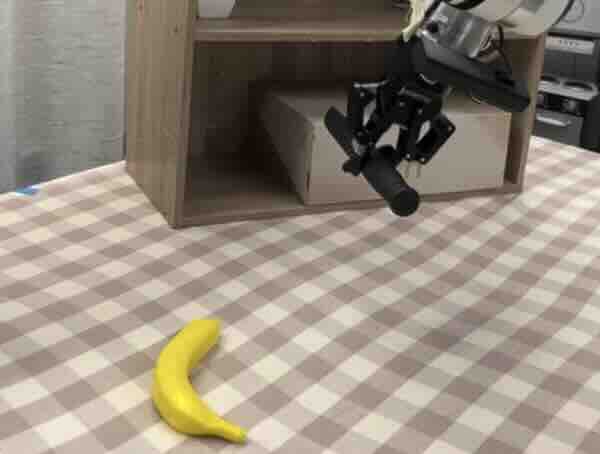} &
    \includegraphics[width=0.095\linewidth]{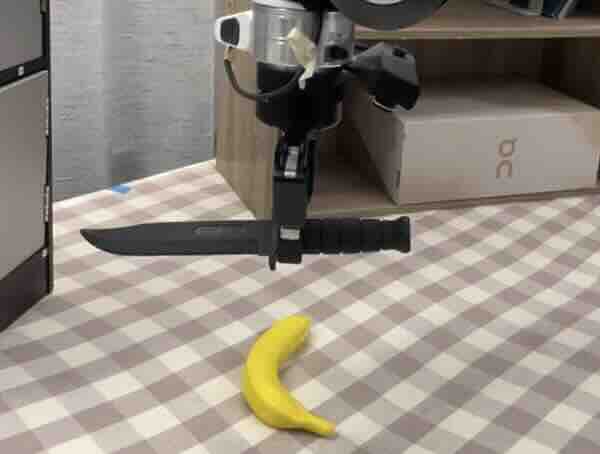} &
    \includegraphics[width=0.095\linewidth]{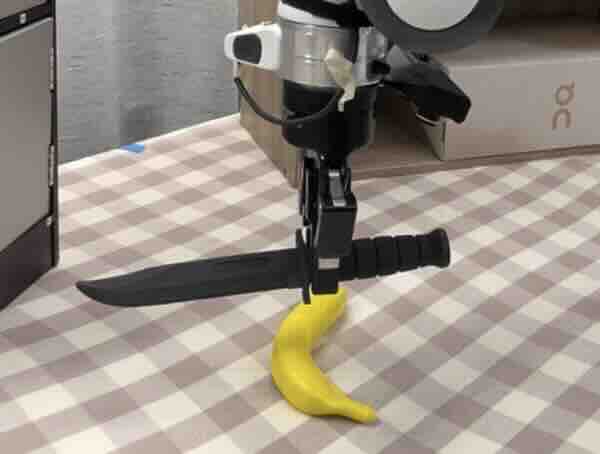} &
    \includegraphics[width=0.095\linewidth]{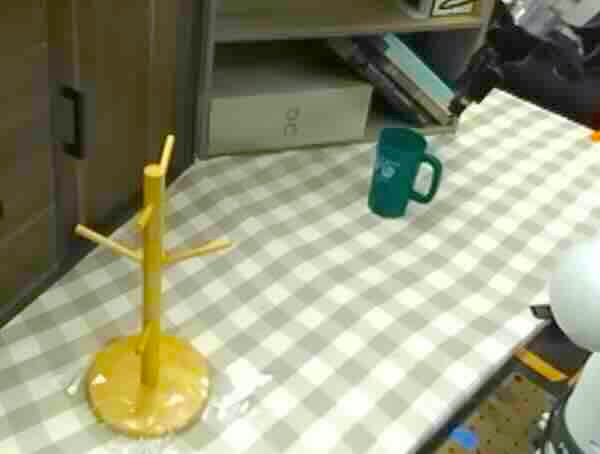} &
    \includegraphics[width=0.095\linewidth]{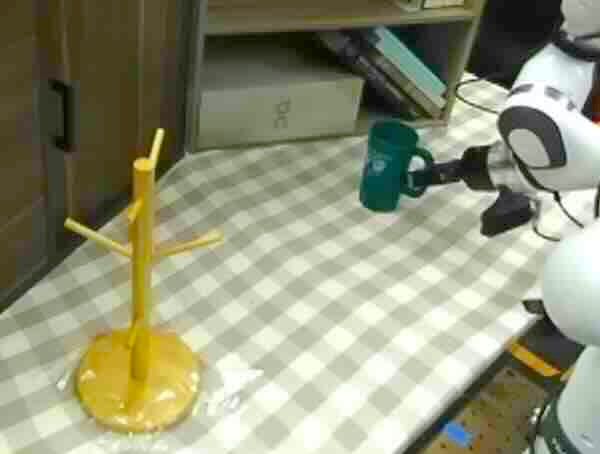} &
    \includegraphics[width=0.095\linewidth]{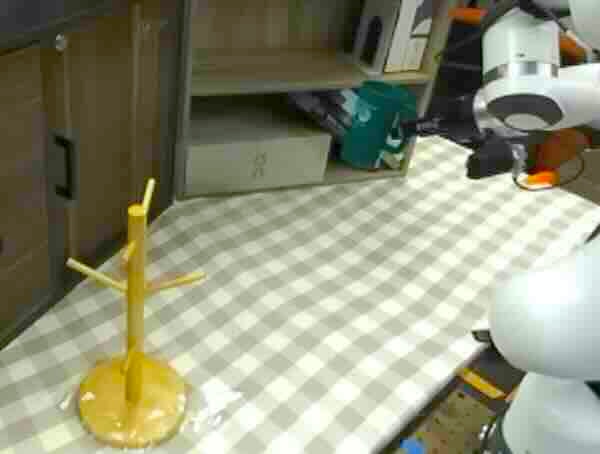} &
    \includegraphics[width=0.095\linewidth]{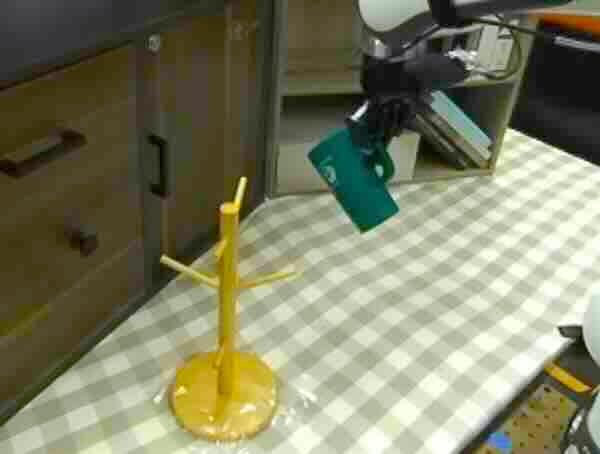} &
    \includegraphics[width=0.095\linewidth]{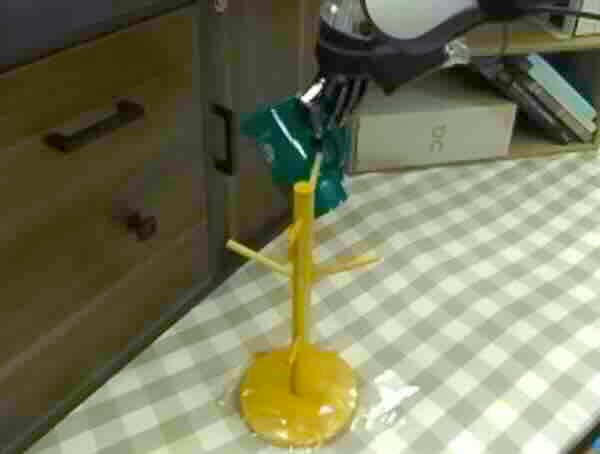} \\
    \multicolumn{5}{l}{\parbox[t]{0.47\linewidth}{\footnotesize\raggedright\textbf{Long-horizon \& semantic:} clean table, largest first}} &
    \multicolumn{5}{l}{\parbox[t]{0.47\linewidth}{\footnotesize\raggedright\textbf{Memory:} erase order instructions, then stack cubes}} \\
    \includegraphics[width=0.095\linewidth]{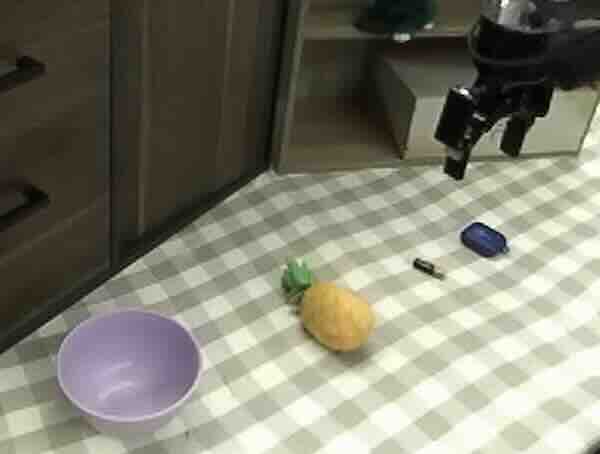} &
    \includegraphics[width=0.095\linewidth]{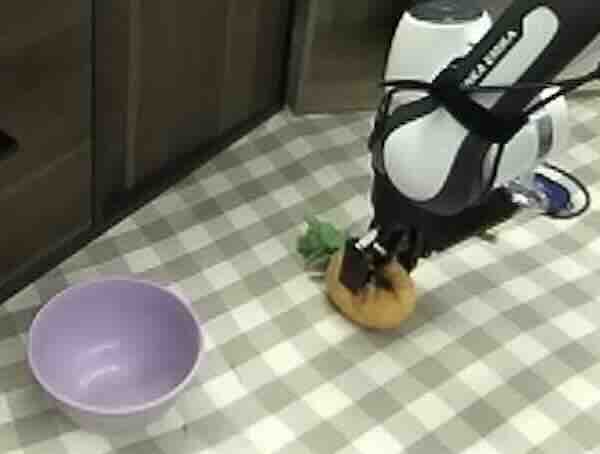} &
    \includegraphics[width=0.095\linewidth]{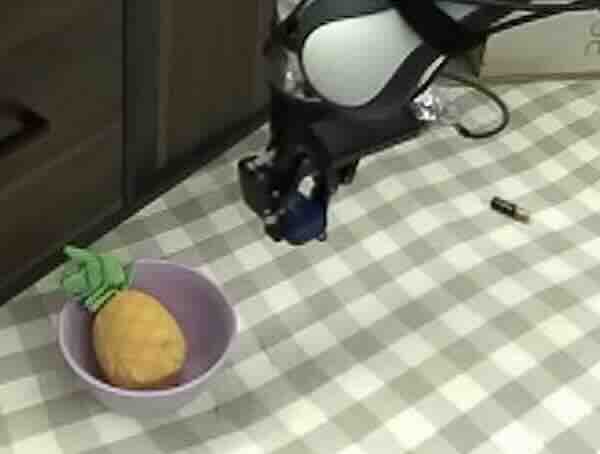} &
    \includegraphics[width=0.095\linewidth]{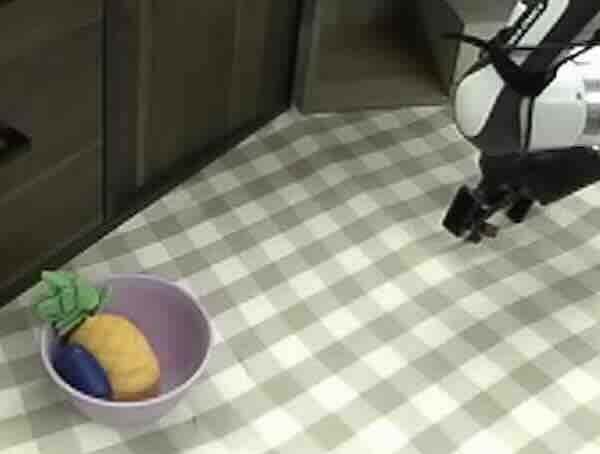} &
    \includegraphics[width=0.095\linewidth]{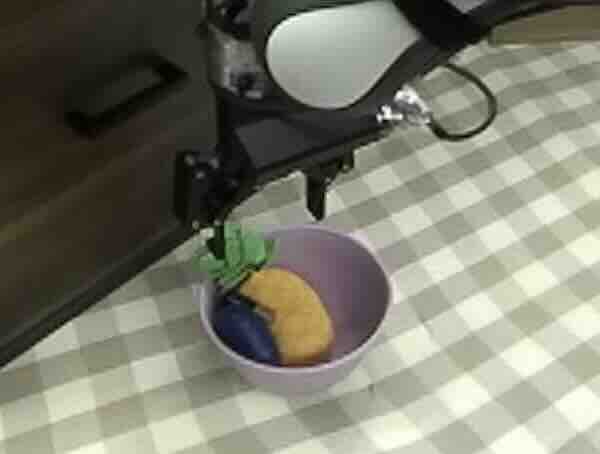} &
    \includegraphics[width=0.095\linewidth]{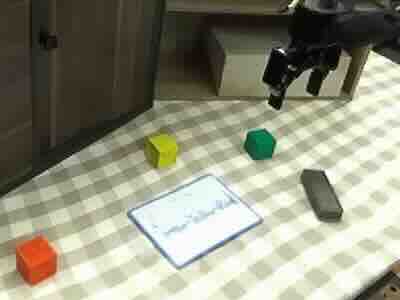} &
    \includegraphics[width=0.095\linewidth]{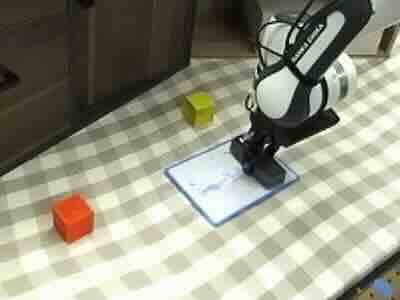} &
    \includegraphics[width=0.095\linewidth]{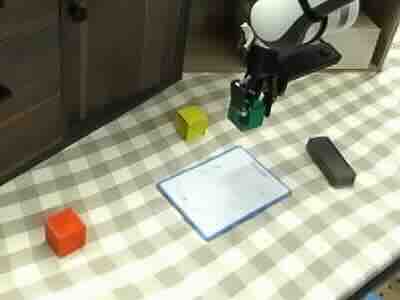} &
    \includegraphics[width=0.095\linewidth]{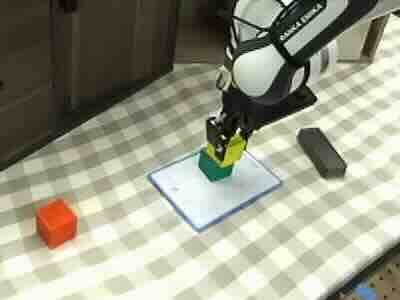} &
    \includegraphics[width=0.095\linewidth]{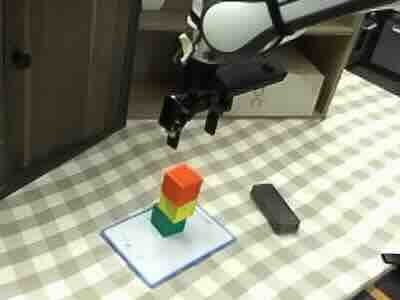} \\
  \end{tabular}

\caption{\textbf{Tabletop manipulation evaluation tasks.} See the \href{https://maestro-robot.github.io/}{project website} for videos.}
  \space \space
\label{fig:tabletop_tasks}
  \vspace{1mm}
\end{figure}

\subsection{Real-World Experiment Setup}
\label{sec:experiment_setup}

We evaluate \methodname{} in the real world along three axes, manipulation-challenge category, evaluation-setting diversity, and robot embodiment, using Gemini Robotics-ER 1.5~\cite{geminirobotics15} as its VLM. For our zero-shot results and tool/harness ablations (Secs.~\ref{sec:tabletop_results}--\ref{sec:scaling_ablation}), we evaluate \methodname{} without invoking the $\pi_{0.5}$ VLA tool. In App.~\ref{app:vla_tool}, we evaluate $\pi_{0.5}$ as a callable tool within \methodname{}.

\textbf{Embodiments.} We evaluate \methodname{} on two embodiments: a Franka-based DROID tabletop platform~\cite{khazatsky2024droid}, which best supports comparison against prior generalist policies (tabletop being the most widely studied domain), and a Unitree Go2-W wheeled quadruped with a top-mounted AgileX PiPER arm for mobile manipulation. Full hardware and camera configurations are in App.~\ref{app:setup_details}.

\textbf{Tasks.} We identify the key challenge axes facing today's generalist robot policies and evaluate \methodname{} on eight tabletop and four mobile manipulation tasks. The eight tabletop tasks span pick-place, deformable object, articulated object, spatial reasoning, tool use, object affordance, long-horizon semantic, and memory challenges (Fig.~\ref{fig:tabletop_tasks}); the four mobile tasks span long-horizon manipulation, long-horizon loco-manipulation, active exploration, and object-affordance reasoning (Fig.~\ref{fig:mobile_tasks}). Each figure visualizes every task with its instruction.

\textbf{Evaluation Protocol.} We adopt the STAR-Gen taxonomy of generalization~\cite{2025stargen}. For each task, we evaluate five settings held fixed across methods: one initial setting and four settings generated using STAR-Gen, with perturbations to task-relevant object appearance, object poses, action verbs, and object identity. These are distinct settings rather than repeated runs of an identical scene. The tabletop comparison comprises eight tasks and four methods, for $40$ scored trials per method and $160$ trials in total. Following $\pi_{0}$~\cite{black2024pi_0}, we score task \emph{progress} per trial for quantitative results (App.~\ref{app:eval_metric}).

 Human reviewers assigned all task-progress and task-completion scores independently of the agent's monitor.

\subsection{Zero-Shot Tabletop Manipulation Results}
\label{sec:tabletop_results}

\begin{figure}[!htbp]
  \centering

  \includegraphics[width=\textwidth]{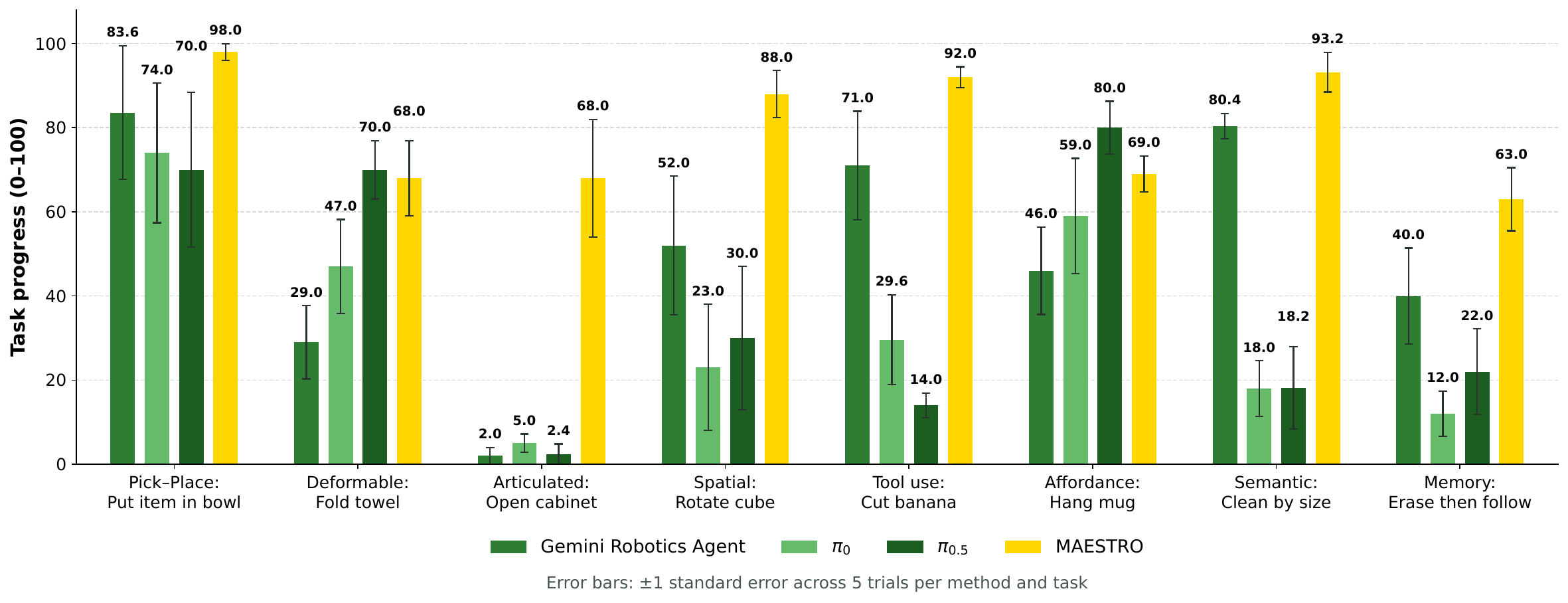}
  \caption{Tabletop manipulation results: mean task progress (0--100; higher is better). Error bars show $\pm1$ standard error across five trials per method and task; numerical values are reported in Table~\ref{tab:tabletop_results}.}
  \space
\label{fig:main-table}
  \vspace{-2mm}
\end{figure}

\noindent \textbf{Baselines.}
We benchmark \methodname against open-source generalist robot policies from both the Code-as-Policies (CaP) and VLA paradigms.%
 \space
For CaP,
we adopt the approach described in the Gemini Robotics technical report
~\cite{team2025gemini}, which uses Gemini to enable zero-shot robot control via code generation. Our baseline implementation follows the system-prompt structure, three-tool inventory (top-down grasp prediction, 2D bounding-box detection, 2D pointing, all performed by the VLM itself rather than via external robotics models), and intra-episode replan loop documented in~\cite[Fig.~12, App.~B.3.2]{team2025gemini}. This corresponds to the ``Shared CaP pattern'' column of Table~\ref{tab:comparison} (App.~\ref{app:method_details}).

We implement this using Gemini Robotics-ER 1.5
~\cite{geminirobotics15} and denote it as the \textbf{Gemini Robotics Agent}.
For VLAs, we compare against \textbf{$\pi_{0}$}
~\cite{black2024pi_0} and \textbf{$\pi_{0.5}$}~\cite{pi2025pi05}. Specifically, we use the $\pi_{0}$-FAST-DROID checkpoint for $\pi_{0}$ ($\pi_{0}$-FAST fine-tuned on the DROID dataset) and the $\pi_{0.5}$-DROID checkpoint for $\pi_{0.5}$, which is similarly fine-tuned.

\noindent\textbf{Results.} Fig.~\ref{fig:main-table} summarizes the tabletop results, with detailed numbers in App.~\ref{app:detailed_tabletop}. \methodname{} has the highest mean task progress on six of eight tasks in the selected five-trial comparison. $\pi_{0.5}$ has the highest mean on fold towel and hang mug on mug holder. VLA baselines struggle on several perturbed settings and lack an explicit memory mechanism. Their mean progress is low on open cabinet; on rotate cube, the selected trials yield means of $23.0$ for $\pi_0$ and $30.0$ for $\pi_{0.5}$, compared with $88.0$ for \methodname{}. The Gemini Robotics Agent, our CaP baseline, plans well but struggles to turn high-level plans into low-level actions: its agentic loop is built around Gemini's own perception~\cite[p.~11]{team2025gemini}, so it is limited to whatever bounding boxes, points, and top-down grasps the VLM produces on its own, and cannot, for instance, localize a cabinet handle without active perception or grasp and orient an eraser or towel corners without external segmentation, depth, and grasp modules.

\methodname{} bridges this gap less through the tool modules than through the orchestration harness around them (Sec.~\ref{sec:monitor}): it hides error-prone chains like point-cloud reconstruction and grasp prediction behind single calls (encapsulation), injects procedural guidance on module choice and exact geometry for reorientation, and renders tool outputs back for the VLM to verify points and 6-DoF poses. When a step fails, its closed-loop plan-react-replan cycle re-perceives and recomposes tools, e.g., recovering from a failed top-down grasp of a shelved tennis ball via active perception and the grasp model for a safer, reachable pose. This lets \methodname{} surpass prior CaP systems in dexterity and task coverage on tasks typically considered better suited to VLA-style approaches, though hang mug on mug holder still exceeds its affordance and orientation reasoning. Rollout videos for all eight tabletop tasks are on our \href{https://maestro-robot.github.io/}{project site}.

\subsection{Zero-Shot Mobile Manipulation Results}
\label{sec:mobile_results}

\begin{figure}[!htbp]
  \centering
  \space
\begin{minipage}[t]{0.56\textwidth}
    \vspace{6pt}
    \centering
    \space
\setlength{\tabcolsep}{1pt}
    \renewcommand{\arraystretch}{0.55}
    \begin{tabular}{*{5}{c}}
      \multicolumn{5}{l}{\parbox[t]{0.96\linewidth}{\footnotesize\raggedright\textbf{Long-horizon manipulation:} collect all toys on table}} \\
      \includegraphics[width=0.18\linewidth]{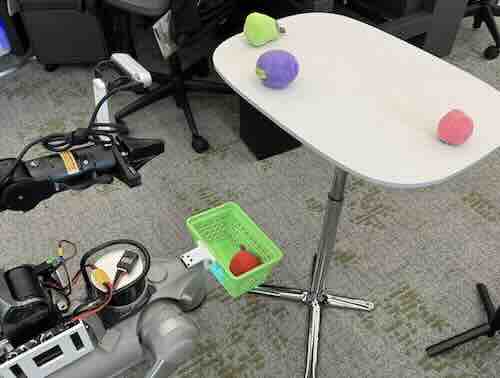} &
      \includegraphics[width=0.18\linewidth]{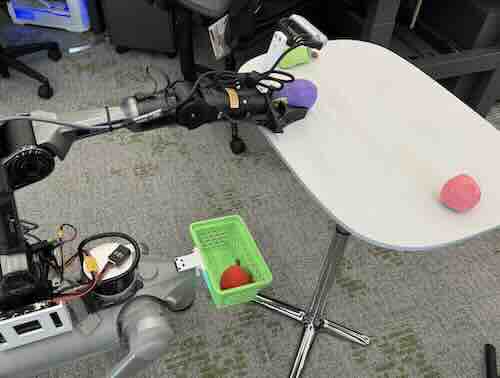} &
      \includegraphics[width=0.18\linewidth]{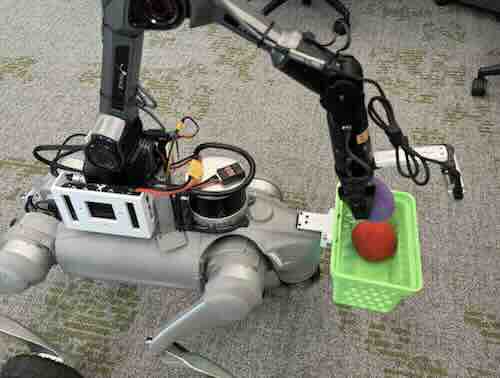} &
      \includegraphics[width=0.18\linewidth]{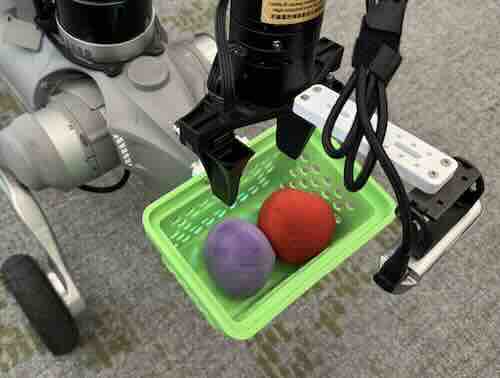} &
      \includegraphics[width=0.18\linewidth]{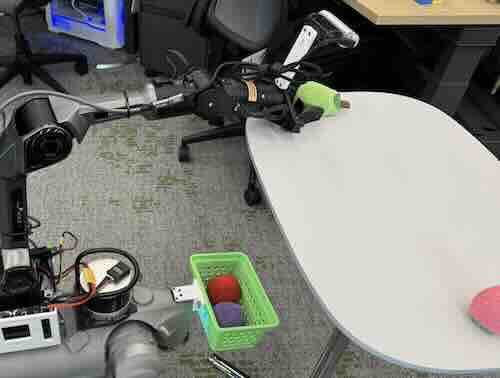} \\
      \multicolumn{5}{l}{\parbox[t]{0.96\linewidth}{\footnotesize\raggedright\textbf{Loco-manipulation:} throw ball into garbage can}} \\
      \includegraphics[width=0.18\linewidth]{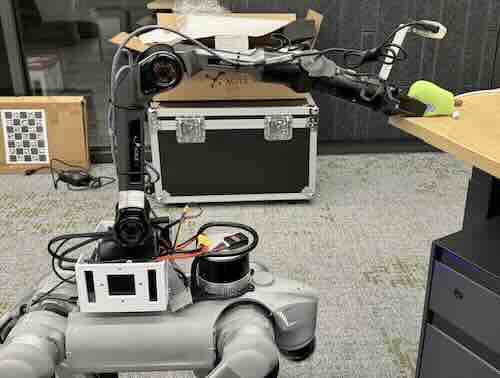} &
      \includegraphics[width=0.18\linewidth]{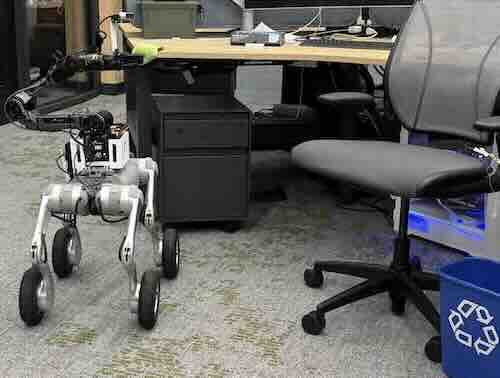} &
      \includegraphics[width=0.18\linewidth]{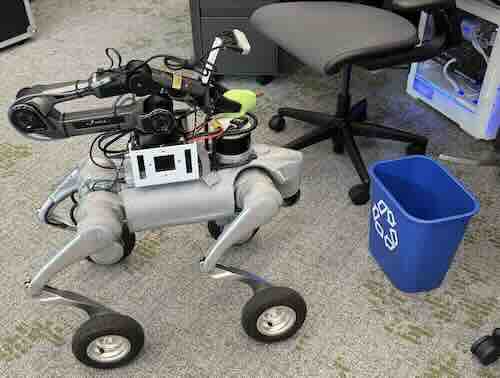} &
      \includegraphics[width=0.18\linewidth]{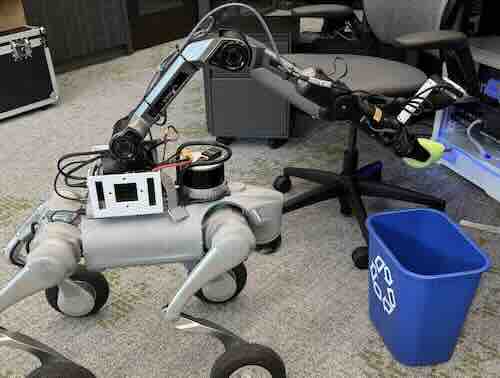} &
      \includegraphics[width=0.18\linewidth]{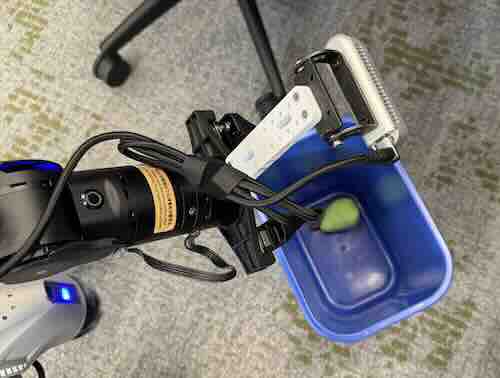} \\
      \multicolumn{5}{l}{\parbox[t]{0.96\linewidth}{\footnotesize\raggedright\textbf{Active exploration:} search for object and return}} \\
      \includegraphics[width=0.18\linewidth]{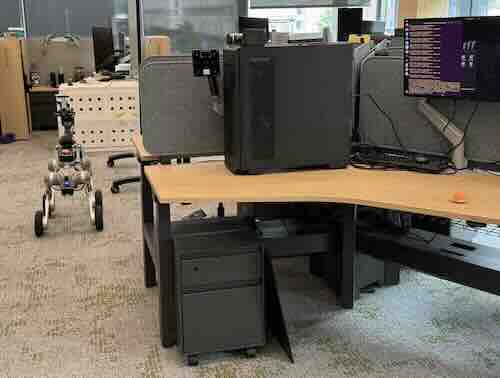} &
      \includegraphics[width=0.18\linewidth]{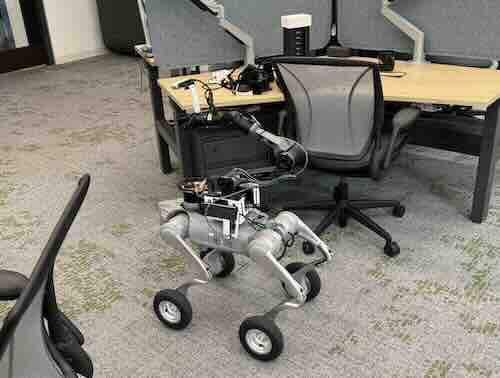} &
      \includegraphics[width=0.18\linewidth]{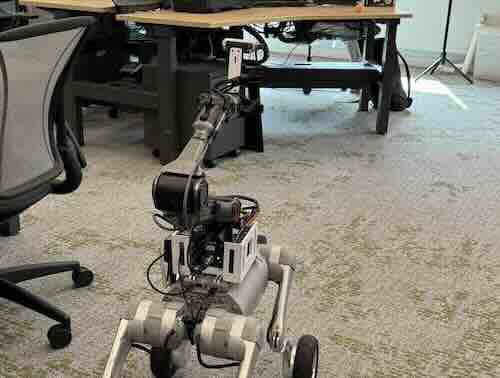} &
      \includegraphics[width=0.18\linewidth]{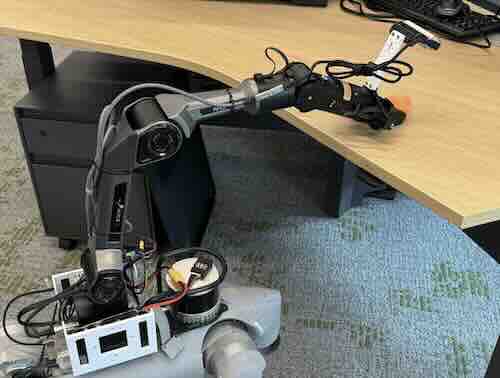} &
      \includegraphics[width=0.18\linewidth]{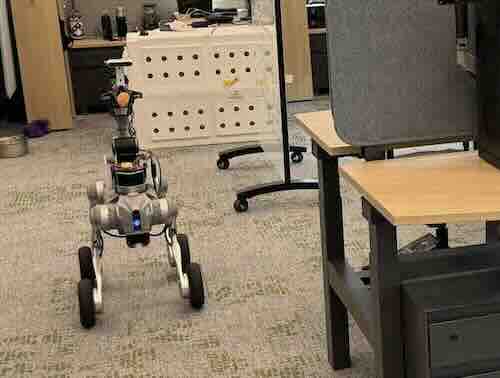} \\
      \multicolumn{5}{l}{\parbox[t]{0.96\linewidth}{\footnotesize\raggedright\textbf{Object affordance:} press button to open door}} \\
      \includegraphics[width=0.18\linewidth]{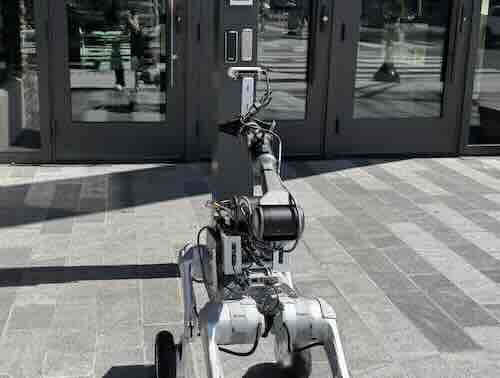} &
      \includegraphics[width=0.18\linewidth]{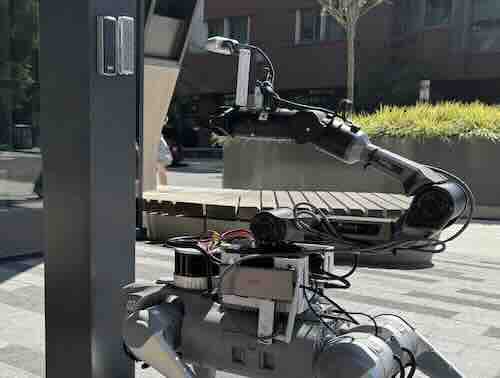} &
      \includegraphics[width=0.18\linewidth]{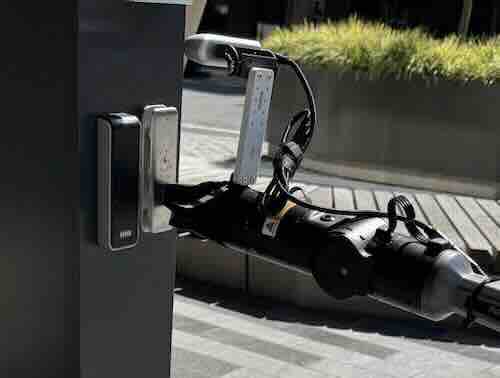} \\
    \end{tabular}

\captionof{figure}{\textbf{Mobile manipulation evaluation tasks.} See the \href{https://maestro-robot.github.io/}{project website} for videos.}
    \space
\label{fig:mobile_tasks}
  \end{minipage}\hfill
  \space
\begin{minipage}[t]{0.42\textwidth}
    \vspace{6pt}
    \centering

    \includegraphics[width=\linewidth]{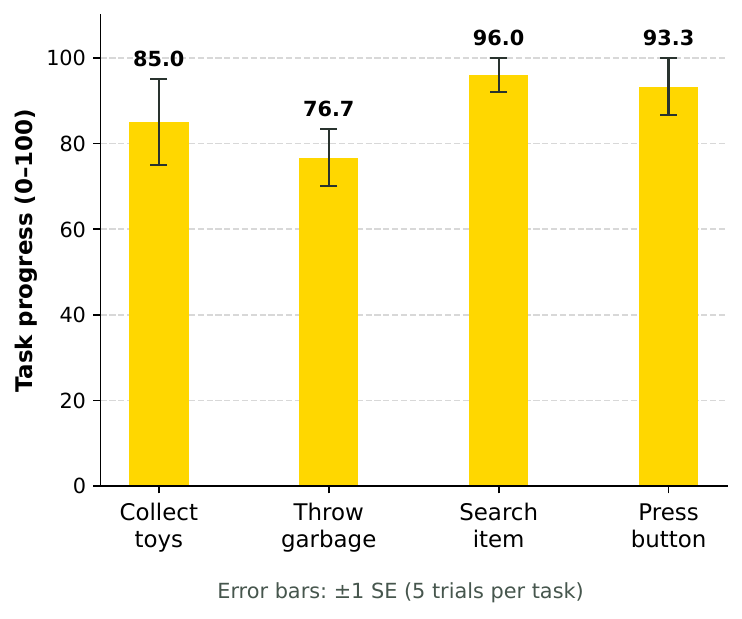}
    \captionof{figure}{Mobile manipulation results: mean task progress (0--100; higher is better). Error bars show $\pm1$ standard error across five trials per task; numerical values are reported in Table~\ref{tab:mobile_results}.}
    \space
\label{fig:mobile_main}
  \end{minipage}
\end{figure}

\noindent \textbf{Baselines.} No open-source zero-shot CaP or VLA generalist policies exist for our wheeled-quadruped embodiment, so we evaluate \methodname{} alone, reporting per-task progress and failure modes.

\setlength{\intextsep}{12pt plus 2pt minus 2pt}%
\noindent \textbf{Results.}
Fig.~\ref{fig:mobile_main} reports mean mobile manipulation progress and $\pm1$ standard error across five trials per task; Table~\ref{tab:mobile_results} gives the numerical means and SEs (App.~\ref{app:detailed_tabletop}).
 \space
The two \textit{long-horizon} tasks show lower progress rates due to their multi-stage object interactions; a cached semantic map substantially improves performance by letting the agent re-track objects and complete sub-tasks without redundant search. \textit{Active exploration} achieves high progress because we provide multi-view images at replan sessions to support spatial reasoning, and \textit{press button} benefits from precise keypoint selection. Remaining failures primarily stem from low-level execution; we decompose them by module class in App.~\ref{app:failure_cases}. Rollout videos for all four mobile tasks are available on our \href{https://maestro-robot.github.io/}{project site}.

\noindent We analyze \methodname{}'s failure modes in detail, broken down per embodiment and module class with concrete trial examples, in App.~\ref{app:failure_cases}.

\Needspace*{14\baselineskip}
\subsection{How important is each component of \methodname?}
\label{sec:scaling_ablation}

Table~\ref{tab:comparison} (App.~\ref{app:method_details}) lists the tools and harness components that \methodname{} adds over Base CaP. We examine their contributions by expanding the tool library from Base CaP (S0) through advanced perception (S1), grasping and control (S2), and geometry and spatial reasoning (S3). At each stage we compare four conditions: \textbf{Full harness (I+E)}, \textbf{Interface only (I)}, \textbf{Execution only (E)}, and \textbf{No harness}. Following the five harness principles defined in Sec.~3.3, Interface retains encapsulation and procedural knowledge, while Execution retains closed-loop control, infrastructure, and feedback. We group the latter three because they work together during and after execution. We report per-task progress for opening a cabinet, rotating a cube so a target face points up, and grasping a knife by its handle to cut a banana. Aggregate values are unweighted means of these three task means.

\begin{figure}[!htbp]
  \centering
  \includegraphics[width=\textwidth]{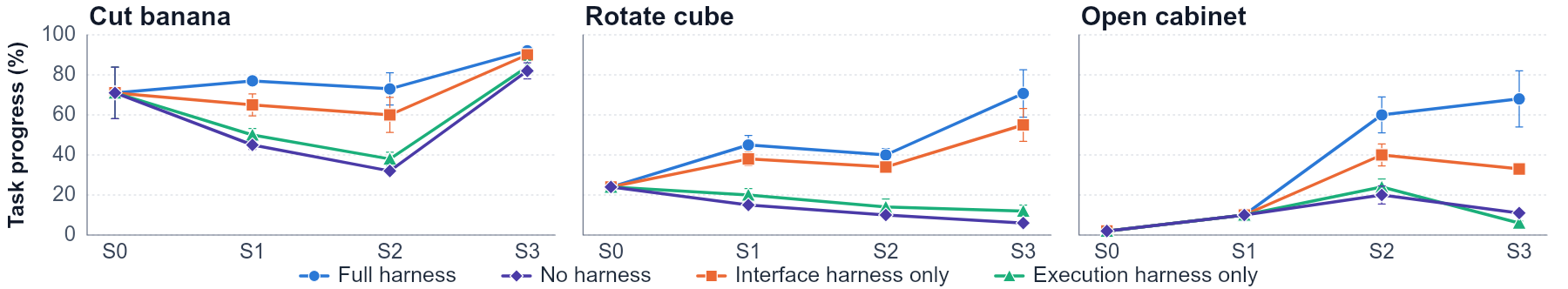}
  \caption{\textbf{Scaling \methodname{}'s tools and harness together.} Mean task progress for cut banana, rotate cube, and open cabinet as tool groups are added to Base CaP (S0): S1 (+perception), S2 (+grasp/control), and S3 (+geometry). The four conditions are Full harness (I+E), Interface only (I), Execution only (E), and No harness. Error bars show per-task standard errors, computed as sample SD divided by $\sqrt{n}$ for each task--stage--condition cell. S0 uses the same five baseline trials per task for all four curves.}
  \label{fig:scaling}
\end{figure}

\textbf{Results.} Fig.~\ref{fig:scaling} shows that the full harness has the highest or tied mean in all nine S1--S3 task-stage comparisons. Its three-task average rises from 32.3\% at S0 to 44.0\% at S1, 57.7\% at S2, and 76.9\% at S3, although individual tasks need not improve at every stage. At S3, Interface only reaches 59.3\%, Execution only 34.0\%, and No harness 33.0\%. The partial-harness variants generally fall between full and no harness, but Execution only is lower than No harness for open cabinet at S3 (6\% versus 11\%). These results support the value of the full harness in this setting and show that adding tools alone does not reliably improve progress.

Complementary analyses corroborate this: per-module \emph{leave-one-out} ablations (App.~\ref{app:loo_ablation}) provide additional module-level comparisons, and a tool-call-level analysis (App.~\ref{app:tooluse}) shows \methodname{} composes task-specific rather than fixed workflows, with 47 of 63 initially failed subtask instances later receiving a logged success judgment (74.6\%). This conditional, log-derived recovery statistic is distinct from human-scored trial completion. The scaling study evaluates S0--S3; the contribution of the S4 VLA tool is evaluated separately in App.~\ref{app:vla_tool}.

\subsection{Beyond single-trial execution}
\label{sec:beyond_single_trial}

\noindent Beyond single-trial execution, we examine three capabilities in preliminary case studies. \methodname{} can refine its generated policies across trials: on one selected open-cabinet trajectory, three rounds of code refinement improve progress from $35\%$ to full completion (App.~\ref{app:evolution}). It can combine task-level planning with faster VLA execution: on table cleanup, invoking $\pi_{0.5}$ as a tool raises mean progress from $89\%$ to $94\%$ and reduces mean completion time by $2.3\times$ (App.~\ref{app:vla_tool}). It can also be used for automatic data generation: starting from 10 \methodname{} execution rollouts, Tether~\cite{liang2026tether} generates a larger dataset across new scene configurations. On two tasks, this pipeline performs comparably to Tether initialized with human-teleoperated demonstrations (App.~\ref{app:tether}).

\section{Limitations and Future Work}
\space
\label{sec:limitations}

\methodname{}'s closed-loop harness catches and recovers from individual tool failures rather than assuming every module always succeeds, which is much of why it outperforms a naive composition of the same tools. Recovery has limits, however: when a module is systematically wrong in a way \methodname{} cannot observe or route around, the error propagates to the task. These residual failures concentrate in perception and low-level control, and more delicate control will require richer low-level behaviors than the current toolset supports.
\space
Runtime remains a limitation (App.~\ref{app:failure_cases}): querying a VLM during planning and replanning makes \methodname{} slower than end-to-end VLAs such as $\pi_{0.5}$. Recent direct-control evaluations of GPT-6 Astra show that inference can take much longer than the robot motion it specifies~\cite{galbot2026astra}. Delegating low-level control to tools avoids generating every action with the VLM, but repeated planning and verification still incur latency and token charges. Future work will explore how to reduce VLM calls while preserving timely failure detection, and how to distill frequently repeated behaviors into faster policies.

\section{Conclusion} \label{conclusion}

We present \methodname, a VLM-driven coding agent for tabletop and mobile manipulation, equipped with a curated tool library and the orchestration infrastructure to compose and adapt these tools. Without collecting new teleoperation data, it outperforms the evaluated VLAs and Base CaP system in our tabletop comparison. Ablations highlight the importance of the harness: adding tools alone does not reliably improve performance. Its modular interfaces allow improved tools and VLMs to be incorporated as they become available, offering a complementary route to scaling robot data.

\acknowledgments{This work was supported in part by DARPA TIAMAT HR0011249042, NSF CAREER 2239301, ONR N00014-22-1-2677, Penn Engineering Futures Fund,  NSF SLES 2331783, and NAIRR Pilot 240077. PC acknowledges a DURIP grant from the Office of Naval Research.}

\raggedbottom
\clearpage
\space
\bibliography{references_arxiv_final}
\clearpage
\appendix
\section*{Appendix}
\section{Additional Method Details}\label{app:method_details}

\textbf{Plan--react--replan execution and safety.} \methodname{}'s generated code runs against a constrained, curated API rather than free-form actuation. Execution errors and planning failures are returned to the agent in \texttt{stdout} for replanning; cuRobo checks modeled collisions and inverse-kinematics feasibility. An interactive replay on a representative trial, including a case where \methodname{} detects mid-execution that the whiteboard was not fully erased and autonomously triggers a recovery, is available on the Agent Dashboard at our \href{https://maestro-robot.github.io/}{project site}.

\noindent\textbf{Implementation.} %
\methodname{}'s code is available at \href{https://github.com/penn-pal-lab/maestro}{github.com/penn-pal-lab/maestro}
.

\subsection{Tool library by scaling stage (S0--S4)}\label{app:toollib_stages}

This subsection gives the staged description of \methodname{}'s tabletop tool library summarized in Sec.~\ref{sec:modules}, presented as a cumulative progression over the Base-CaP starting point that mirrors the scaling study of Sec.~\ref{sec:scaling_ablation}: each stage replaces a capability the Base-CaP VLM handles with its default abilities by the curated, best-quality module for that capability. Table~\ref{tab:comparison}(a) lists the same library compactly, and Sec.~\ref{sec:franka_modules} and Sec.~\ref{sec:mobile_modules} give the full per-module mechanics.

\textbf{S0 (Base CaP): the VLM's default abilities.} The starting point exposes only what the VLM can do by default (Sec.~\ref{sec:background}): 2D pointing and bounding boxes for localization, VLM-imagined top-down grasps, and basic Cartesian and gripper control. Each cluster below replaces one of these defaults with the best-quality module available for that capability.

\textbf{S1 (+advanced perception).} We replace the VLM's default perception with a \emph{coarse-to-fine} hierarchy of specialized vision modules that lets the agent trade speed against precision: raw RGB and proprioception (fastest, coarsest); Grounded-SAM segmentation-mask centroids and 2D pointing (medium); FoundationStereo~\cite{wen2025foundationstereo} metric depth, which supplies the reliable 3D positions and dense object point clouds that the VLM's monocular guesses cannot; and VLM-selected task-relevant keypoints (ReKep-inspired~\cite{huang2024rekep}) for precise but slower localization of features a VLM struggles to point at directly. Two harness capabilities, \emph{active perception} and \emph{image-editing feedback}, make this hierarchy usable in practice; we detail them with the rest of the perception harness in Sec.~\ref{sec:monitor}. None of these abilities exist in Base CaP, which perceives only as well as the generalist VLM itself, with no metric depth and no point clouds.

\textbf{S2 (+grasping and control).} Base CaP can only execute the top-down grasps the VLM imagines and moves the arm by direct Cartesian servoing with no notion of obstacles; we replace both. For grasping, GraspGen~\cite{murali2025graspgen} is a learned grasp model that predicts ranked 6-DoF grasps on an object's point cloud, expressing the side, angled, and functional-part grasps a top-down pose cannot. Because grasp quality hinges on a clean point cloud, the grasp pipeline first drives a planned multi-view scan of the wrist camera and fuses the views into a single object point cloud before prediction. For control, we add cuRobo~\cite{curobo_report23} point-cloud-based collision-free motion planning, letting \methodname{} reach those grasps and move through cluttered scenes while avoiding unintended contact, which is essential whenever the arm must make and break contact with objects or operate around obstacles. Both capabilities are entirely absent at S0--S1, where the agent has neither a 6-DoF grasp source nor any collision awareness.

\textbf{S3 (+geometry and spatial reasoning).} In our initial experiments, the agent struggled with metric spatial reasoning even with improved perception. We therefore provide explicit tools for computing distances, vectors, and rotations from perception outputs. These tools are not included in the Base CaP baseline.

\textbf{S4 (+VLA).} Finally, \methodname{} treats a learned visuomotor policy, $\pi_{0.5}$~\cite{pi2025pi05}, as a callable tool for fast, reliable execution of short skills such as pick-and-place. A raw VLA has no self-termination signal, so exposing it as a usable tool depends on the monitoring harness of Sec.~\ref{sec:monitor}; the agent selects the VLA only when it fits the subtask (App.~\ref{sec:vla_tool}) and falls back on the procedural modules above otherwise.

\textbf{Mobile manipulation.} For the mobile embodiment, the same library is extended with embodiment-specific modules: mobile-base state estimation, active-perception tools (look left / right / ground, view the carry-on basket, log object locations), navigation, a fine-grained ``nudge'' tool for local base adjustment, and a semantic map that caches observed object locations to support persistent reasoning over long-horizon tasks.

\subsection{Per-stage harness details}\label{app:harness_details}

This subsection gives the full mechanics of the per-cluster harness summarized in Sec.~\ref{sec:monitor}, one paragraph per scaling stage; Table~\ref{tab:comparison}(b) lists the same infrastructure compactly, and the tags (\hform{a})--(\hform{e}) refer to the five harness design principles of Sec.~\ref{sec:monitor}.

\textbf{S1 perception harness.} The perception harness adds: priority guidance on which localizer to try first and when to escalate from pointing to segmentation to keypoints, plus a ``detect all objects once and cache'' rule that avoids repeating an ${\sim}30$\,s scan (\hform{b},~\hform{d}); active perception that lets the agent reposition the wrist camera, zooming in or looking around, so the downstream segmentation, depth, and keypoint tools receive informative observations, since off-the-shelf vision models are only as good as the views they are given (\hform{d}); image-editing feedback that draws predicted points and 6-DoF poses back onto the observation image, so the VLM can visually confirm a result it cannot read as raw numbers (\hform{e}); and an encapsulated keypoint tool that hides a multi-step pipeline behind one call: best-view selection, segmentation, candidate-point sampling, image rendering, depth projection, and a second VLM that picks the matching point (\hform{a}).

\textbf{S2 grasp and control harness.}
For grasping, \emph{encapsulation} (\hform{a}) packages a multi-step tool chain behind a single call.
 \space
A single \texttt{get\_grasp\_pose} call internally runs active perception (a planned spherical look-around of the wrist camera with multi-view fusion into a clean object point cloud), then segmentation, depth, and GraspGen, returning a ranked 6-DoF grasp; a companion \texttt{plan\_grasp} fuses that grasp with control, applying a gripper-flange calibration offset and collision-free planning before executing (\hform{a}, plus calibration knowledge the VLM should not manage). Prompt-level guidance adds a grasp sanity-check guardrail and side-vs-top-down strategy (\hform{b}). Because these modules are heavy, the harness runs them on a GPU server, handling file sync and client--server communication, and a scene-scan cache persists multi-view captures for reuse (\hform{d}). The agent simply asks for a grasp.

\textbf{S3 geometry harness.} The geometry primitives are already atomic, so their harness is purely procedural (\hform{b}): in-context worked examples of composing vectors and rotations (e.g.\ building an axis from two keypoints and rotating about it) and guidance on when to use them for orientation, so the agent reaches for exact geometry instead of guessing poses.

\textbf{S4 VLA harness.} $\pi_{0.5}$ runs fast but has no signal for when it has finished. \methodname{} wraps it in \texttt{execute\_with\_pi05},
which runs the policy alongside a separate Qwen-2.5-VL monitor
 \space
that checks task-relevant completion conditions on the live image at ${\sim}2$\,Hz and interrupts the policy on success or when a replan is needed (\hform{a},~\hform{c}).
A resource manager coordinates exclusive camera access and handles starting and stopping policy execution, allowing the VLA to run alongside the main coding agent (\hform{d}).
 \space
This monitor is precisely what turns a non-terminating policy into a tool the agent can call and reliably stop.

\textbf{Curating for quality, not for external models per se.} The best-quality modules \methodname{} adopts are usually specialized, externally-developed systems, but that is a consequence of curating for quality rather than a goal in itself, since the generalist VLM is rarely the best tool for a precise robotics subtask.

\textbf{Mobile-manipulation replanning.} In mobile manipulation settings, before performing failure analysis and rewriting code, \methodname{} is also prompted to actively look around and perceive the environment to build a more complete understanding of its surroundings before it replans.

\begin{table}[H] %
  \centering
  \caption{Comparison of \methodname's tool library and harness against \emph{Base CaP}, the closed-loop coding-agent pattern \methodname{} builds on (Sec.~\ref{sec:background};
the Gemini Robotics agent~\cite[Fig.~12, App.~B.3.2]{team2025gemini} is a documented example
). Components in \textbf{bold} are introduced or substantially scaled in \methodname; cells marked \underline{\textit{None}} indicate no equivalent in Base CaP. Both parts are organized by the scaling stage, from S0 (Base CaP) through S4, at which each capability enters: part (a) lists the curated robotics modules added at each stage (Sec.~\ref{sec:modules}), and part (b) lists the harness infrastructure each stage requires (Sec.~\ref{sec:monitor}). Tags (\hform{a})--(\hform{e}) in part (b) name the five harness design principles of Sec.~\ref{sec:monitor}.}
  \label{tab:comparison}
  \vspace{3pt} %
  \setlength{\tabcolsep}{4pt}
  \renewcommand{\arraystretch}{1.2}
  \rowcolors{3}{gray!15}{white}
  \begin{small}
  \resizebox{\textwidth}{!}{%
  \begin{tabular}{>{\raggedright\arraybackslash}p{3.6cm}
                  >{\raggedright\arraybackslash}p{4.6cm}
                  >{\raggedright\arraybackslash}p{7.3cm}}
    \toprule
    \textbf{Stage / component} & \textbf{Base CaP} & \textbf{\methodname} \\
    \midrule
    \multicolumn{3}{l}{\textit{(a) Tool library --- curated robotics modules composed as callable tools, by scaling stage (Sec.~\ref{sec:modules})}} \\
    \midrule
    S0 (Base CaP) & VLM-internal 2D pointing, bounding boxes, imagined top-down grasps; Cartesian + gripper control~\cite{team2025gemini,geminirobotics15} & \textit{Starting point; each row below replaces a default with a curated module} \\
    \textbf{S1: advanced perception} & VLM performs perception itself~\cite{Zawalski24-ecot,duan2024manipulateanything,team2025gemini} & \textbf{Grounded-SAM segmentation centers; FoundationStereo~\cite{wen2025foundationstereo} metric depth; ReKep-inspired task-relevant keypoints~\cite{huang2024rekep}; active perception (zoom / look around with wrist camera); point and 6-DoF-pose image-editing feedback}\textsuperscript{$\dagger$} \\
    \textbf{S2: grasping \& control} & Imagined top-down grasp only; direct Cartesian moves, no collision awareness~\cite{team2025gemini,geminirobotics15} & \textbf{GraspGen~\cite{murali2025graspgen} 6-DoF grasps (side / angled / functional-part) with multi-view fusion; cuRobo~\cite{curobo_report23} collision-free motion planning} \\
    \textbf{S3: geometry \& spatial reasoning} & \underline{\textit{None}} (language-only) & \textbf{Distance, vector construction, rotation between vectors, rotation about an axis} \\
    \textbf{S4: VLA} & \underline{\textit{None}} & \textbf{$\pi_{0.5}$~\cite{pi2025pi05} VLA with high-frequency closed-loop monitoring (Qwen-2.5-VL)} \\
    \midrule
    \multicolumn{3}{l}{\textit{Additional modules for mobile manipulation}} \\
    Perception (mobile) & Build global/local map~\cite{zhi2025comerobot} & Mobile-base state estimation; \textbf{active perception (look left / right / ground, view carry-on basket, log object location)} \\
    Locomotion (mobile) & Navigation~\cite{zhi2025comerobot} & Navigation; \textbf{fine-grained ``nudge'' tool for local adjustment} \\
    \midrule
    \multicolumn{3}{l}{\textit{(b) Harness --- orchestration that makes the tools usable, grouped by stage (Sec.~\ref{sec:monitor})}} \\
    \midrule
    Closed-loop plan-react-replan (shared substrate, c) & \checkmark\ Intra-episode, stdout + image after each code block & \checkmark\ Intra-episode, monitor-driven per-step revision \\
    Code execution + stdout feedback & \checkmark & \checkmark \\
    \textbf{S1: perception harness} & \underline{\textit{None}} & \textbf{Pointing/segmentation/keypoint priority + detect-once caching (\hform{b},\,\hform{d}); point and 6D-pose feedback drawn into the observation (\hform{e}); encapsulated keypoint tool (\hform{a})} \\
    \textbf{S2: grasp \& control harness} & \underline{\textit{None}} & \textbf{\texttt{get\_grasp\_pose}/\texttt{plan\_grasp} tools bundling active perception, segmentation, depth, GraspGen and cuRobo (\hform{a}); grasp guardrail + side/top-down guidance (\hform{b}); GPU client--server + scan cache (\hform{d})} \\
    \textbf{S3: geometry harness} & \underline{\textit{None}} & \textbf{In-context vector/rotation composition examples + orientation guidance (\hform{b})} \\
    \textbf{S4: VLA harness} & \underline{\textit{None}} & \textbf{\texttt{execute\_with\_pi05} + Qwen-2.5-VL ${\sim}2$\,Hz monitor interrupting on completion/replan (\hform{a},\,\hform{c});
resource manager for camera handoff and starting and stopping policy execution (\hform{d})
} \\
    \textbf{Cross-trial reflection / code evolution} & \underline{\textit{None}} (intra-episode only) & \textbf{Per-trial video-grounded success/failure analysis fed in-context to the next trial (c)} \\
    \textbf{
Embodiments evaluated here
} &
Franka tabletop
 \space
& \textbf{Franka tabletop + mobile manipulator} \\
    \bottomrule
  \end{tabular}
  }%
  \end{small}
  \vspace{1mm}\\
  \footnotesize\textsuperscript{$\dagger$}Concurrent with this work, Gemini Robotics-ER 1.6~\cite{geminirobotics16} has begun adding Python-based image manipulation for ambiguity resolution.
\end{table}%

\begin{figure}[t]
  \centering
  \includegraphics[width=0.5\textwidth]{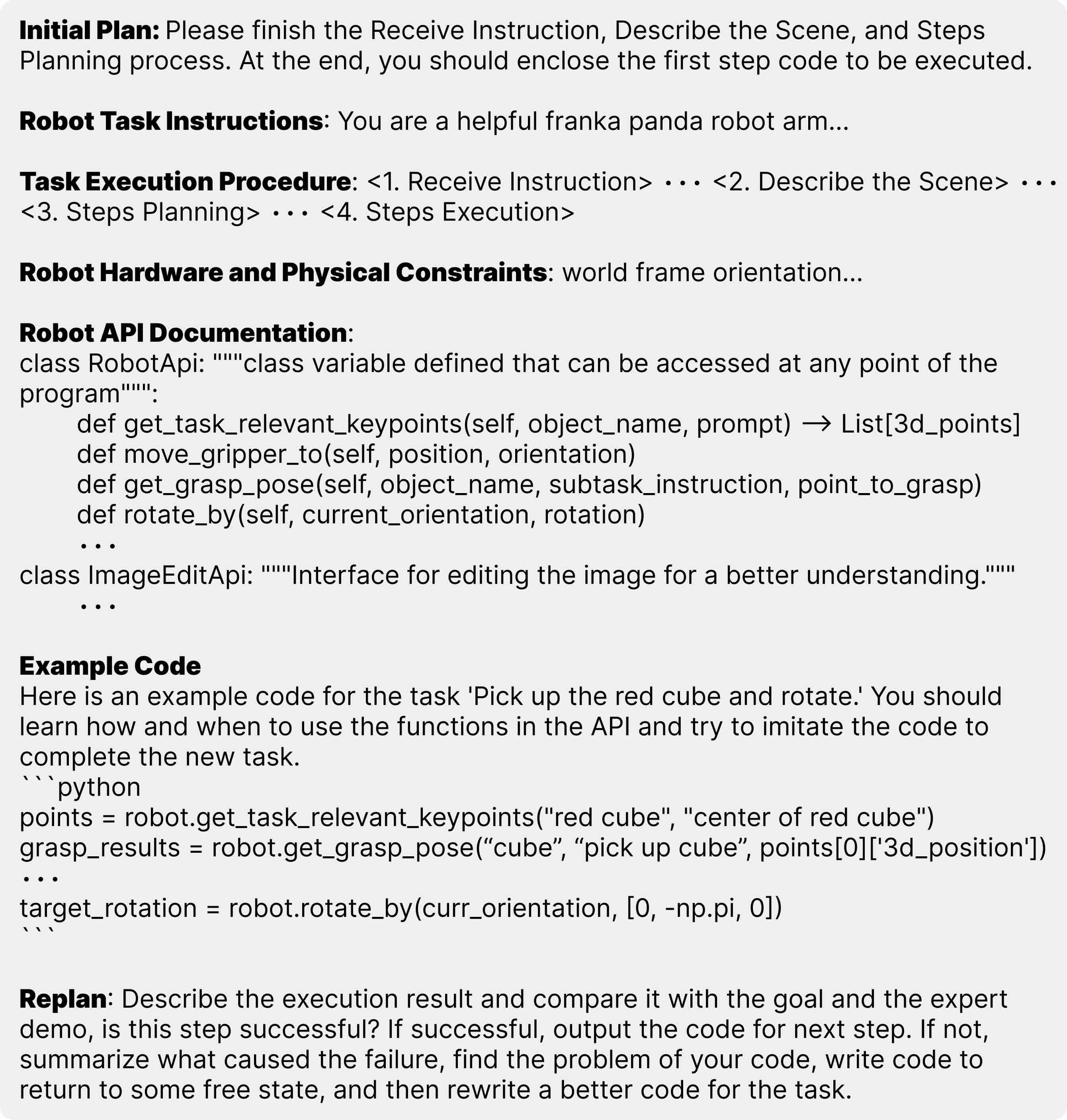}
  \caption{A summarized overview of \methodname's system prompt.}
  \space
\label{fig:prompt}
  \vspace{1mm}
\end{figure}

\begin{figure}[t]
  \centering
  \includegraphics[width=0.7\linewidth]{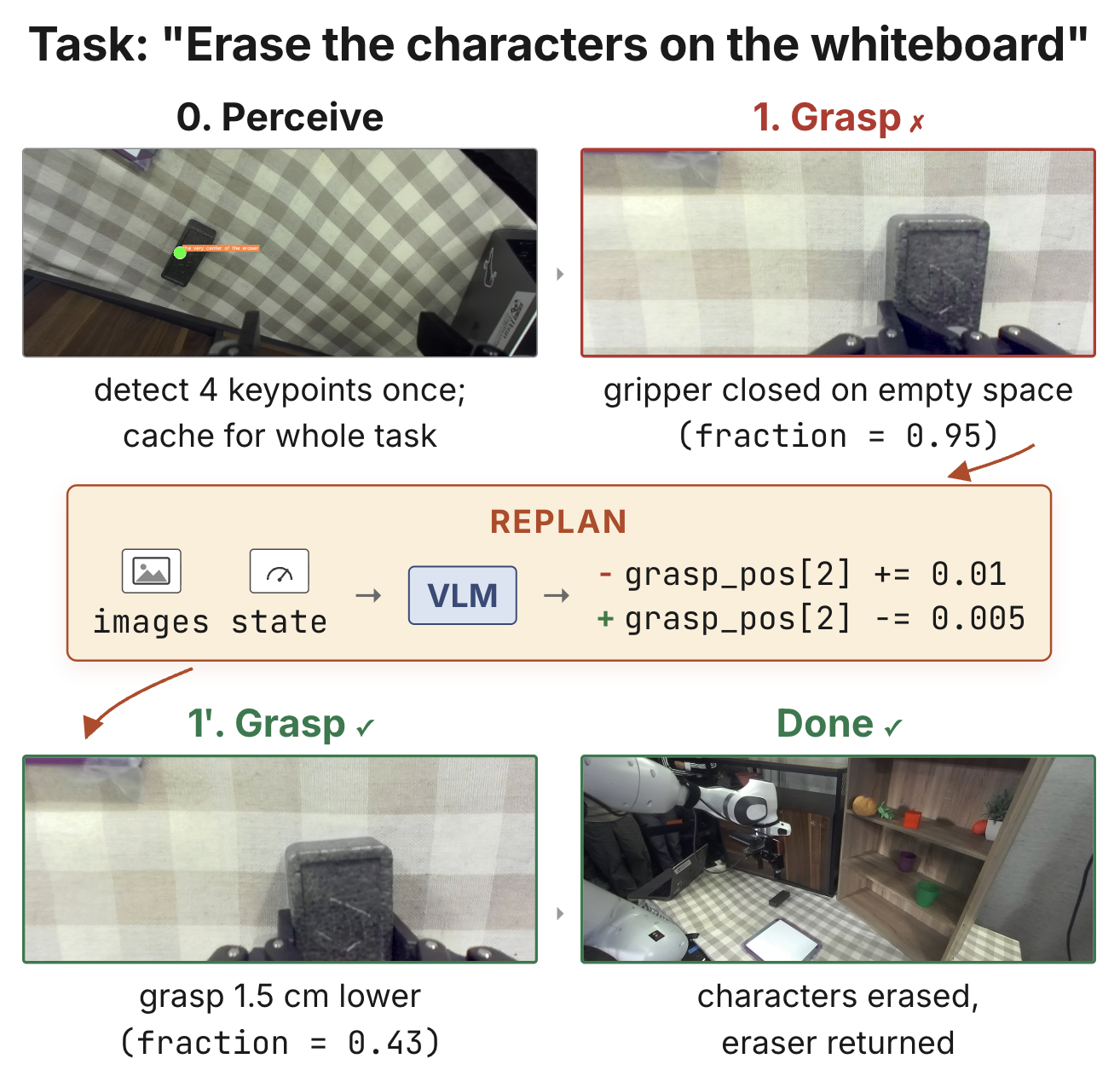}

\caption{One replan iteration on ``Erase the characters on the whiteboard''. The first \texttt{actor\_code} grasps 1\,cm too high (top); from the post-execution image and robot state, \textbf{\textit{replan}} rewrites a single line (\texttt{grasp\_pos[2]}) and the retry succeeds, completing the task with no further replans (bottom). See the \href{https://maestro-robot.github.io/}{project website} for a video of this replanning example.}
  \space
\label{fig:replan}
\end{figure}

\begin{algorithm}[t]
\caption{\methodname execution loop (single trial). The cross-trial evolution loop that wraps this procedure is given in Algorithm~\ref{alg:maestro_evolution}.}
\label{alg:maestro}
\begin{algorithmic}[1]
\Require Task $\mathcal{T}$, initial image $I_0$, tool library $\mathcal{M}$, system prompt $\mathcal{P}_{\text{sys}}$, VLM $\pi$, substep budget $S$
\Statex \textit{Symbols:} $c$ (actor code generated by the VLM), $o$ (stdout from running $c$), $I$ (post-execution image), $r$ (post-execution robot state), $h$ (within-trial trace), $\sigma$ (react status flag, e.g.\ \textsc{TaskComplete}).
\State $h \gets [\,]$
\State $c \gets \pi_{\text{plan}}(\mathcal{P}_{\text{sys}}, \mathcal{T}, I_0)$ \Comment{plan task; emit code for the first substep}
\For{substep $k = 1, \dots, S$}
  \State $(o, I, r) \gets \textsc{Execute}(c;\, \mathcal{M})$ \Comment{run substep code to completion}
  \State append $(c, o, I, r)$ to $h$
  \State $(\sigma, c) \gets \pi_{\text{react}}(\mathcal{T}, c, o, I, r, h)$ \Comment{Sec.~\ref{sec:monitor}: judge substep, then emit next-substep code or replanned code}
  \If{$\sigma = \textsc{TaskComplete}$} \textbf{break} \EndIf
\EndFor
\end{algorithmic}
\space
\end{algorithm}

\begin{algorithm}[t]
\caption{\methodname execution and evolution loop (full version). The inner plan-react-replan loop is the single-trial Algorithm~\ref{alg:maestro} (above); the outer \texttt{for} loop over trials is the cross-trial evolution mechanism of Sec.~\ref{sec:evolution}.}
\label{alg:maestro_evolution}
\begin{algorithmic}[1]
\Require Task $\mathcal{T}$, initial image $I_0$, tool library $\mathcal{M}$, system prompt $\mathcal{P}_{\text{sys}}$, VLM $\pi$, max trials $N$, per-trial substep budget $S$
\Statex \textit{Symbols:} $c$ (actor code generated by the VLM), $o$ (stdout from running $c$), $I$ (post-execution image), $r$ (post-execution robot state), $h_n$ (trial-$n$ trace), $\mathcal{H}$ (across-trial history), $\sigma$ (react status flag, e.g.\ \textsc{TaskComplete}).
\State $\mathcal{H} \gets \emptyset$ \Comment{across-trial history (Sec.~\ref{sec:evolution})}
\For{trial $n = 1, \dots, N$}
  \State $h_n \gets [\,]$
  \State $c \gets \pi_{\text{plan}}(\mathcal{P}_{\text{sys}}, \mathcal{T}, I_0, \mathcal{H})$ \Comment{plan task; emit code for the first substep}
  \For{substep $k = 1, \dots, S$}
    \State $(o, I, r) \gets \textsc{Execute}(c;\, \mathcal{M})$ \Comment{run substep code to completion}
    \State append $(c, o, I, r)$ to $h_n$
    \State $(\sigma, c) \gets \pi_{\text{react}}(\mathcal{T}, c, o, I, r, h_n)$ \Comment{Sec.~\ref{sec:monitor}: judge substep, then emit next-substep code or replanned code}
    \If{$\sigma = \textsc{TaskComplete}$} \textbf{break} \EndIf
  \EndFor
  \State $\mathcal{H} \gets \mathcal{H} \cup \{h_n\}$ \Comment{persist trace as in-context history for next trial (Sec.~\ref{sec:evolution})}
\EndFor
\end{algorithmic}
\space
\end{algorithm}

\subsection{Tabletop Manipulation Modules}
\label{sec:franka_modules}
We equip our tabletop manipulation agent with a suite of tools spanning perception, reasoning, control, pre-trained visuomotor policies, and image editing. The tools are provided in \textbf{bolded text} below.

\textbf{Perception.} For perception tools, we adopt a ``coarse-to-fine'' approach. At the fastest and simplest level, the agent can access \textbf{raw sensory inputs} (RGB images and robot proprioception). Because raw depth is often noisy, we use FoundationStereo~\cite{wen2025foundationstereo} to estimate more reliable \textbf{depth maps} from RGB. For \textbf{object center point}, given a language description, we employ Grounded-SAM to produce a mask and its center point, while a Gemini-based pointing tool returns single 2D points corresponding to language queries. For tasks requiring higher precision---such as grasping towel corners---we provide a \textbf{salient task-relevant points} tool inspired by ReKep~\cite{huang2024rekep}: after generating a segmentation mask, we overlay a uniform point grid and ask GPT-o3 to select the most relevant points. This tool hierarchy balances granularity and runtime: raw images are coarse but fast, while salient points are slower but more precise. \methodname autonomously reasons about which tools to use and how to use them for each task, achieving a balance between execution speed and task performance. As an alternative, we also provide \textbf{language-based pointing} using Gemini Robotics-ER 1.5~\cite{geminirobotics15}.

\textbf{Reasoning.} In our initial experiments, \methodname{} struggled with task-relevant spatial reasoning even with advanced perception tools. We therefore provide tools for explicit spatial calculations: \textbf{measuring distances}, \textbf{constructing vectors using points}, \textbf{computing relative rotations between vectors}, and \textbf{rotating a vector by specified angles}.

\textbf{Control.} For low-level manipulation, we provide Cartesian end-effector control tools: \textbf{move gripper to}, \textbf{open gripper}, and \textbf{close gripper}. We use cuRobo~\cite{curobo_report23} for point-cloud-based motion planning with collision checks. At the grasping stage, collision checking is disabled only for the gripper links identified by the robot and gripper URDFs; checks remain active for the rest of the robot. This contact exception does not model the grasped object as part of the robot, so collisions involving held objects remain a limitation (App.~\ref{app:failure_cases}).

\textbf{Learned Visuomotor Policies.} We equip \methodname with two types of learned visuomotor policies: a \textbf{grasp model} and a \textbf{VLA}. For grasping, we provide GraspGen~\cite{murali2025graspgen}, while for general-purpose low-level visuomotor control we provide the state-of-the-art $\pi_{0.5}$ model~\cite{pi2025pi05}. A key challenge when integrating end-to-end VLAs as tools is determining when to interrupt their execution, since they continue running inference until externally stopped. We use a separate completion monitor running at 2 Hz, hosting Qwen-2.5-VL-72B-Instruct locally to check task completion with a yes/no question and interrupt execution when completion is detected.

\textbf{Image Editing.} Prior work~\cite{yang2023setofmark,fangandliu2024moka} shows that adding visual marks can improve the grounding and reasoning abilities of VLMs. Thus, we provide \methodname with image-editing tools that can \textbf{draw points} and \textbf{overlay 6D poses} on images, enabling clearer spatial references and richer visual reasoning during manipulation. Concretely, \methodname draws the 2D points returned by its pointing tools back onto the observation so the VLM can confirm a target selection before acting (e.g.\ verifying which object or contact point was chosen in spatial-reasoning and pick-place tasks), and overlays a candidate 6D grasp pose so the agent can check the gripper's approach direction against object affordances (e.g.\ confirming a knife is grasped by the handle, or a mug at an orientation that permits hanging). The edited image is fed back into the next replanning step, closing the loop on these checks.

\subsection{Additional Modules for Mobile Manipulation}
\label{sec:mobile_modules}
To extend beyond tabletop tasks, we equip \methodname with additional tools that enable mobile manipulation.

\textbf{Perception.} For a mobile manipulator, obtaining the full \textbf{robot 6D state} requires more than raw proprioception. We therefore employ Faster-LIO~\cite{bai2022faster}, a lightweight LiDAR–Inertial Odometry method, to provide robust state estimation of the mobile base even under visually challenging conditions. This enhanced perception ensures reliable navigation and spatial reasoning across larger workspaces. In addition, mobile manipulation benefits from active perception to build a more complete understanding of the surroundings and gather task-relevant information on demand. To this end, we provide \methodname with active perception tools---\textbf{look left}, \textbf{look right}, \textbf{look to the ground}, and \textbf{view carry-on basket}, \textbf{remember object location}---which allow it to scan the environment, adjust its viewpoint, and access the basket contents as needed.

\textbf{Locomotion.} We supply two tools: (1) the \textbf{nudge} tool, which applies small velocity adjustments for precise positioning near a target location, and (2) the \textbf{navigation} tool, which leverages Nav2\cite{macenski2020marathon} to move the robot safely to a target pose on the map. This dual interface allows \methodname to fluidly switch between global navigation and fine-grained local control.

\textbf{Manipulation.} To support transport tasks, we provide a dedicated \textbf{put in basket} tool that enables the robot to efficiently carry multiple objects during long-horizon mobile manipulation. For unloading, \methodname leverages the \textbf{view carry-on basket} perception tool to inspect the basket's contents and then automatically generates the code needed to move objects out of the basket.

\section{Beyond Single-Trial Execution: Evolution, VLA Tool Use, and Autonomous Data Collection}\label{app:beyond_execution}

Beyond the single-trial zero-shot evaluation, this appendix presents three case studies. \textbf{Cross-trial evolution:}\label{sec:evolution_results} from the least successful initial open-cabinet run ($35\%$ progress), three rounds of video-grounded code refinement reach full task completion (App.~\ref{app:evolution}, Fig.~\ref{fig:evolution}); this single selected trajectory illustrates adaptation rather than estimating its average benefit across tasks. \textbf{VLA as a tool:}\label{sec:vla_tool} the default library includes $\pi_{0.5}$~\cite{pi2025pi05} for short pick-and-place skills. On \emph{clean up the table, largest object first}, using it raises average progress from $89\%$ to $94\%$ and reduces wall-clock completion time by $2.3\times$ while retaining \methodname{}'s high-level planning (App.~\ref{app:vla_tool}).
\textbf{Automatic data generation:}\label{sec:tether_exp} on two tasks, Tether~\cite{liang2026tether} starts from 10 \methodname{} execution rollouts and generates a larger dataset across new scene configurations; humans reset the environment between rollouts.
 \space
The reported success rates are comparable to human-teleoperated Tether on these tasks ($95\%$ versus $100\%$ on \emph{Bowl from Shelf}, and $88\%$ on \emph{Wiping with Cloth}); Fig.~\ref{fig:data_collection} and App.~\ref{app:tether} give the comparison.

\section{Evolution Across Trials: Open-Cabinet Case Study}\label{app:evolution}

This appendix expands the evolution result summarized in App.~\ref{sec:evolution_results}, tracing the trial-by-trial code evolution that the outer loop of Algorithm~\ref{alg:maestro_evolution} (App.~\ref{app:method_details}) produces on the open-cabinet task.

\begin{figure*}[htbp] %
  \centering
  \includegraphics[width=0.7\textwidth]{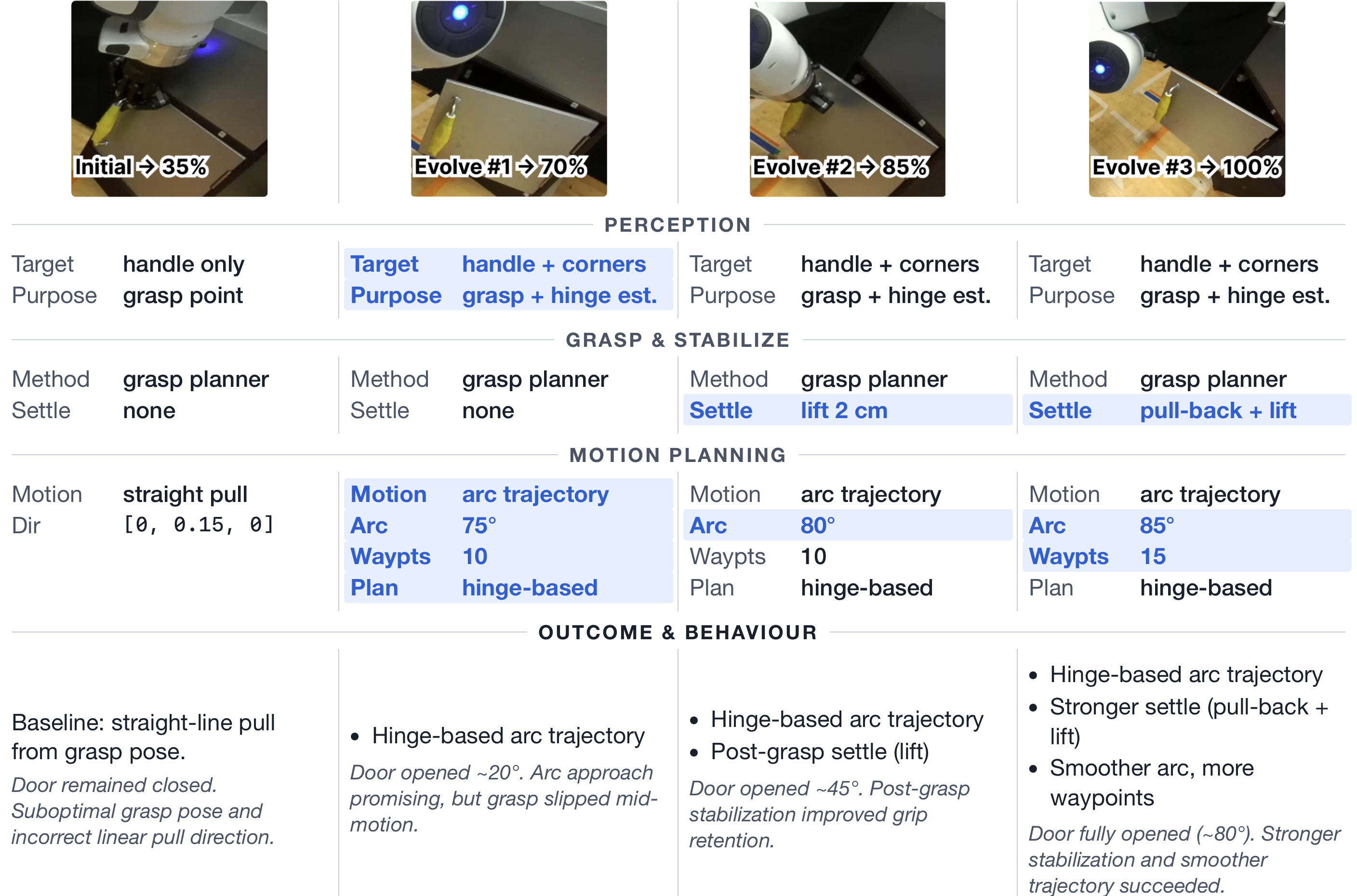}
  \caption{\textbf{Self-evolution of \methodname's generated policy on the ``open cabinet door'' task.} Top: representative frames with completion progress. Middle: semantic code abstraction with changed parts highlighted in blue. Bottom: strategy changes and outcome per trial. Across four autonomous trials, the agent evolves from a naive straight-line pull to a hinge-based arc trajectory with grasp stabilization, achieving full task completion.}
  \label{fig:evolution}
  \vspace{-2mm}
\end{figure*}

We present a case study of \methodname{}'s evolution mechanism (the outer loop of Algorithm~\ref{alg:maestro_evolution}, App.~\ref{app:method_details}) on the open-cabinet task. The trajectory starts from the least successful initial run ($35\%$ progress). Fig.~\ref{fig:evolution} shows the policy abstraction, strategy change, and outcome at each refinement stage.

\textbf{Trial 0 (Initial, 35\%).} \methodname identified the handle but used a top-down grasp and a fixed straight-line pull ([0.0, 0.15, 0.0]) that ignored the hinge kinematics, leaving the door closed. Reflection correctly attributed the failure to the linear motion primitive.

\textbf{Trial 1 (Evolve \#1, 70\%).} \methodname restructured the motion strategy: it added perception of the top-right and bottom-right door corners to estimate the hinge axis, then replaced the straight pull with a hinge-arc trajectory (75°, 10 waypoints). The door began to open, but the handle slipped mid-arc because the grasp could not sustain the rotational forces.

\textbf{Trial 2 (Evolve \#2, 85\%).} To counter the slip, \methodname added a post-grasp stabilization step (a 2 cm upward lift to tighten the grip), widened the arc to 80°, and refined the rotation to incremental orientation updates in the 2D plane. The door reached about 45° before the handle slipped again.

\textbf{Trial 3 (Evolve \#3, 100\%).} \methodname replaced the lift with a combined retract-and-lift (2 cm + 2 cm) that pre-tensions the grasp, widened the arc to 85°, increased waypoints from 10 to 15 (smoother, lower-jerk trajectory), and added incremental yaw tracking so the gripper follows the arc. These changes eliminated grasp slip and the door opened to about 80°.

Across four autonomous trials, \methodname progressed from a linear pull to a kinematically grounded arc policy with robust grasp stabilization, showing that video-grounded self-reflection enables improvements in generated robot programs across trials without human guidance.

\section{Extended Experimental Setup}\label{app:setup_details}

\textbf{Embodiments.} Our tabletop platform follows the DROID setup~\cite{khazatsky2024droid}: a 7-DoF Franka Emika Panda arm with a Robotiq 2F gripper, a wrist-mounted camera, and a third-person camera. Our mobile platform is a Unitree Go2-W wheeled quadruped with a top-mounted AgileX Robotics PiPER manipulator arm and a calibrated wrist-mounted camera for egocentric perception.

The STAR-Gen perturbation protocol and per-task scoring rubric used in evaluation are detailed in App.~\ref{app:eval_metric} below.

\section{Task Evaluation Rubric}
\label{app:eval_metric}

We performed rigorous evaluation on all tasks following STAR-Gen~\cite{2025stargen} to generate new trial for zero-shot generalization capabilities. 5 trials each task (1 initial setup following 4 generated by STAR-Gen), and for each STAR-Gen generated trial, vary all of the following:
\begin{itemize}
    \item Object placement
    \item Object instance
    \item Scene / Lab setting
    \item Language instruction (paraphrase)
\end{itemize}

We use a task-specific progress rubric on a 0--100 scale, assigning credit to verifiable substeps to distinguish how far the agent progressed in incomplete long-horizon trials and identify where it failed. Human reviewers assigned every task-progress and task-completion score independently of the agent's monitor.

The evaluation metrics along with STAR-Gen variations for table-top manipulation tasks are detailed below:
\begin{enumerate}
    \item Pick-place (\textit{``Put item in bowl."}) \\
    \space [25\%] Approach the item. \\
    \space [50\%] Grasp the item. \\
    \space [60\%] Lift up the item. \\
    \space [75\%] Approach the target location. \\
    \space [100\%] Place correctly.

    \item Deformable object (\textit{``Fold the four corners of the towel into the center."}) \\
    For this task,  STAR-Gen introduces significant geometric and interactive variability across trials (e.g., varying object shape and location), we could not rely on one fixed metric. Instead, the task completion rubric was dynamically fine-tuned for each generated trial, ensuring a fair evaluation of progress for sub-goals of each scenario.
    \begin{enumerate}
        \item \textit{``Fold the four corners of the towel into the center."} (color of the towel is different between two trials) \\
        \space [25\%] One corner folded in. \\
        \space [50\%] Two corners folded in. \\
        \space [75\%] Three corners folded in. \\
        \space [100\%] All four corner folded in successfully.

        \item \textit{``Fold the T-shirt into a rectangle."}\\
        \space [15\%] One t-shirt sleeve folded in.\\
        \space [30\%] Both t-shirt sleeves folded in.\\
        \space [60\%] Bottom of t-shirt folded in. \\
        \space [100\%] Fold entire t-shirt inward successfully.

        \item \textit{``Place one corner of the towel to its diagonal corner."}\\
        \space [30\%] Approach one corner.\\
        \space [60\%] Grasp and lift up the corner.\\
        \space [100\%] Place at the diagonal corner successfully.

        \item \textit{``Unfold the towel into a square."}\\
        \space [30\%] Accurately approach the folded corner.\\
        \space [60\%] Grasp and lift up the corner.\\
        \space [100\%] Place at the table to finish unfolding.
    \end{enumerate}

    \item Articulated  object (\textit{``Open cabinet."})\\
    \space [10\%] Approach the cabinet.\\
    \space [40\%] Grasp the handle.\\
    \space [60\%] Attempt to pull.\\
    \space [70\%] Open the door slightly.\\
    \space [100\%] Open the door completely.

    \item Spatial reasoning (\textit{``Rotate cube purple side up."}) \\
    \space [10\%] Approach the cube.\\
    \space [30\%] Grasp the cube.\\
    \space [60\%] Rotate the cube.\\
    \space [100\%] Purple side faces up.

    \item Tool use (\textit{``Cut banana with knife."}) \\
    \space
\space [20\%] Approach the knife.\\
    \space [50\%] Grasp and lift up the knife.\\
    \space [60\%] Approach the banana.\\
    \space [80\%] Position in a good cutting orientation.\\
    \space [100\%] Cut the banana.

    \item Object affordance (\textit{``Hang mug on mug holder."}) \\
    \space [25\%] Approach the mug.\\
    \space [50\%] Grasp and lift up the mug.\\
    \space
\space [75\%] Approach the mug holder.\\
    \space
\space [100\%] Hang on to the stand bar.

    \item Memory \& long-horizon \& semantic (\textit{``Erase instructions on whiteboard, then follow instruction to stack cubes."})\\
    \space
\space [20\%] Pick up the eraser.\\
    \space
\space [30\%] Erase the white board partially.\\
    \space [50\%] Erase the white board completely.\\
    \space [60\%] Return the eraser.\\
    \space [80\%] Stack the second cube on top of the first cube.\\
    \space [100\%] Stack the third cube on top of the second cube.

\end{enumerate}

The evaluation metrics for mobile manipulation tasks are detailed below:

\begin{enumerate}
    \item Long-horizon manipulation (\textit{``Collect all toys on table."}) \\
        \space [25\%] Collect one toy. \\
        \space [50\%] Collect two toys. \\
        \space [75\%] Collect three toys. \\
        \space [100\%] Collect all four toys.

    \item Long-horizon loco-manipulation (\textit{``Throw the ball into garbage can."})\\
    \space [16.7\%] Find the ball.\\
    \space [33.3\%] Move to the ball.\\
    \space [50\%] Pick up the ball.\\
    \space [66.7\%] Find the trash can.\\
    \space [83.4\%] Move to the trash can.\\
    \space [100\%] Drop the ball into the trash can.

    \item Active exploration (\textit{``Search item and return when grasped."}) \\
        \space [20\%] Explore around the area. \\
        \space [40\%] See the object. \\
        \space [60\%] Move to the object. \\
        \space [80\%] Grasp the object. \\
        \space [100\%] Return to the initial position.

    \item Object affordance (\textit{``Press button to open door."})\\
        \space [33.3\%] Identify the correct label. \\
        \space [66.6\%] Approach to the correct button. \\
        \space [100\%] Press button to open the door.
\end{enumerate}

\section{Extended discussion on prior work}
\label{app:extended_related_work}

A wide range of prior work has explored the use of large models --- Vision-Language Models (VLMs) and Large Language Models (LLMs) --- to support specific components of robotic systems. In the common paradigm, these models are embedded into manually designed, typically rigid, modular pipelines, where large models take on one or several well-defined roles while the remaining pipeline is built through ``good old-fashioned engineering'' by humans. These roles span reward and environment design~\cite{ma2023eureka,ma2024dreureka,liang2024eurekaverse, le2024articulate, gao2025vlmgineer}, constraint-function design~\cite{huang2023voxposer,huang2024copa,huang2024rekep,patel2025iker}, high-level planning~\cite{google2022saycan,ha2023scalingup}, self-reflection~\cite{huang2022innermonologue}, visual question answering~\cite{feng2025reflectvlm,fangandliu2024moka,google2024pivot,yuan2024robopoint}, in-context learning~\cite{dipalo2024kat,yin2024roboprompt}, and task-progress evaluation~\cite{liu2023reflect,duan2024aha,ma2024gvl}.

While effective within their intended scope, these approaches remain constrained by their rigid workflows. Because large models automate only a small portion of the pipeline, substantial manual effort is still required to design the rest. This rigidity also limits scalability: systems tuned and fixed for specific settings struggle to generalize to diverse, in-the-wild scenarios. As a result, these methods fall short of the requirements for scalable, general-purpose robotic manipulation.

Rather than reworking the agentic architecture, \methodname builds on the closed-loop VLM coding-agent pattern of recent CaP-based manipulation systems such as Gemini Robotics~\cite{team2025gemini, geminirobotics15, geminirobotics16}. Our contribution is the design of the harness around that shared pattern: rather than relying on the VLM for perception, \methodname composes curated external robotics models as callable tools (Table~\ref{tab:comparison}), letting the system inherit ongoing advances in those subfields without retraining. The result is a system that can autonomously decide \emph{how}, \emph{when}, and \emph{which} modules to employ. In doing so, we show that, alongside the dominant data-scaling route to generalist manipulation, investing in this kind of agentic system is a complementary axis that already yields VLA-level performance without new teleoperation data.

\textbf{Hierarchical and reasoning-augmented VLAs.} Several VLA variants augment the action-generation backbone with language or reasoning while still selecting actions through a fine-tuned VLA: Hi Robot~\cite{shi2025hirobot} decomposes user instructions into natural-language sub-commands that a learned VLA executes, and ECoT-Lite~\cite{chen2025ecot-lite} distills efficient embodied chain-of-thought reasoning directly into the VLA.

\textbf{Concurrent code-as-policies efforts.} Several recent efforts explore approaches related to \methodname{}. CaP-X~\cite{fu2026capx} from NVIDIA proposes a benchmark and framework for code-as-policies coding agents, validated on a Franka arm in simulation and on real hardware, and shares our focus on multi-turn interaction, structured feedback, and skill synthesis. GaP~\cite{chen2026gap} uses a multi-agent coding harness to construct graph policies from modular robot skills and refine them through simulation rehearsal for variational automation. Gemini Robotics 1.5~\cite{geminirobotics15} introduces an ER/VLA stack with generic third-party tool-calling, and Gemini Robotics-ER 1.6~\cite{geminirobotics16} adds Python-based image manipulation for ambiguity resolution. We view this convergence as encouraging evidence that the route is worth pursuing; \methodname{}'s complementary scope is its curated, robotics-specific module library and the orchestration harness built around composing those models in a real-world generalist system across diverse tasks and embodiments.

\section{Detailed Tabletop Manipulation Results}\label{app:detailed_tabletop}

This section provides additional details on the tabletop manipulation results summarized in Fig.~\ref{fig:main-table} of the main text. Table~\ref{tab:tabletop_results} reports the mean and standard error of task progress scores across all five evaluation trials for each task and method.

\textbf{Qualitative analysis.} VLM-based agents excel at high-level semantic reasoning and planning. However, our CaP baseline, the Gemini Robotics Agent, still struggles to turn these high-level plans into effective low-level actions. Its agentic loop is designed around Gemini's own perception capabilities~\cite[p.~11]{team2025gemini}, so on harder tasks the agent is constrained by whatever bounding boxes, points, and top-down grasps the VLM can produce on its own. In open cabinet, this means the agent cannot localize the handle without an active-perception scan, and its top-down grasping strategy no longer applies. Similar limitations show up in fold towel and erase whiteboard, where the agent can plan but cannot localize towel corners or correctly grasp and orient the eraser without external segmentation, depth, and grasp-prediction modules in the loop.

\methodname{} bridges this gap by coupling VLM-level semantic reasoning with a broad suite of specialized tool modules. It leverages grasp models for reliable pick-and-place, active perception modules for accurate point-cloud reconstruction of cabinet handles, task-relevant keypoints for towel corner localization, and geometric reasoning to rotate objects to precise orientations (e.g., turning a cube to place a target color face upward). When errors occur, \methodname{}'s ``react and replan'' mechanism analyzes execution history and images, then revises its code and module orchestration. For example, to pick up the tennis ball on the shelf, it first attempts a top-down grasp but, after detecting rolling and collision risks, switches to active perception to refine the point cloud and leverages the grasp model tool to generate a safer, reachable grasp pose. Nevertheless, the ``hang mug on mug holder'' task reveals \methodname{}'s current limitations: when complex chain-of-thought reasoning about object affordances and spatial orientation is required, it can still struggle. In this case, it fails to correctly align the mug handle's hole with the branch orientation of the holder.

\begin{table*}[h]
  \centering
  \caption{Tabletop manipulation results: mean $\pm$ standard error of task progress (0--100; higher is better) across methods and tasks.}
  \label{tab:tabletop_results}
  \setlength{\tabcolsep}{6pt}
  \renewcommand{\arraystretch}{1.2}
  \rowcolors{2}{gray!15}{white}
  \begin{small}
  \resizebox{\textwidth}{!}{%
  \begin{tabular}{>{\raggedright\arraybackslash}p{2.5cm}
                  >{\raggedright\arraybackslash}p{3.5cm}
                  cccc}
    \toprule
    \textbf{Challenge} & \textbf{Task Description} &
    \makecell{\textbf{Gemini}\\\textbf{Robotics Agent}} &
    \makecell{$\pi_{0}$} &
    \makecell{$\pi_{0.5}$} &
    \makecell{\textbf{\methodname}} \\
    \midrule

    Pick–Place                      & Put item in bowl & $83.6 \pm 15.9$ & $74.0 \pm 16.6$ & $70.0 \pm 18.4$ & $\mathbf{98.0 \pm 2.0}$ \\
    Deformable object                     & Fold the four corners of the towel into the center & $29.0 \pm 8.7$ & $47.0 \pm 11.2$ & $\mathbf{70.0 \pm 6.9}$ & $68.0 \pm 8.9$ \\
    Articulated  object                    & Open cabinet & $2.0 \pm 2.0$ & $5.0 \pm 2.2$ & $2.4 \pm 2.4$ & $\mathbf{68.0 \pm 14.0}$ \\
    Spatial reasoning    & Rotate cube purple side up & $52.0 \pm 16.5$ & $23.0 \pm 15.0$ & $30.0 \pm 17.0$ & $\mathbf{88.0 \pm 5.6}$ \\
    Tool use & Cut banana with knife & $71.0 \pm 12.9$ & $29.6 \pm 10.7$ & $14.0 \pm 2.9$ & $\mathbf{92.0 \pm 2.5}$ \\
    Object affordance & Hang mug on mug holder & $46.0 \pm 10.4$ & $59.0 \pm 13.7$ & $\mathbf{80.0 \pm 6.3}$ & $69.0 \pm 4.3$ \\
    {Long-horizon \& semantic} & {Clean up the table, largest object first} & $80.4 \pm 3.0$ & $18.0 \pm 6.6$ & $18.2 \pm 9.8$ & $\mathbf{93.2 \pm 4.7}$ \\
    Memory \& long-horizon \& semantic                          & Erase instructions on whiteboard, then follow instruction to stack cubes & $40.0 \pm 11.4$ & $12.0 \pm 5.4$ & $22.0 \pm 10.2$ & $\mathbf{63.0 \pm 7.5}$ \\
    \space
\bottomrule
  \end{tabular}
  }%
\end{small}
\end{table*}

\begin{table}[t]
  \centering
  \caption{Mobile manipulation results: mean $\pm$ standard error (SE) of task progress (0--100; higher is better) for \methodname{}, with $n=5$ trials per task. SEs are computed from the reported rounded SDs as $\mathrm{SE}=\mathrm{SD}/\sqrt{5}$.}
  \label{tab:mobile_results}
  \setlength{\tabcolsep}{3pt}
  \renewcommand{\arraystretch}{1.2}
  \rowcolors{2}{gray!15}{white} %
  \begin{small}
  \begin{tabular}{>{\raggedright\arraybackslash}p{2.8cm}
                  >{\raggedright\arraybackslash}p{3.7cm}
                  c}
    \toprule
    \textbf{Task Category} & \textbf{Task Description} & \makecell{\textbf{\methodname}} \\
    \midrule
    Long-horizon manipulation & Collect all toys on table & $85.0 \pm 10.0$ \\
    Long-horizon loco-manipulation & Throw green ball into garbage can & $76.7 \pm 6.7$ \\
    \space
Active exploration & Search for item and return & $96.0 \pm 4.0$ \\
    \space
Object affordance & Press button to open door & $93.3 \pm 6.7$ \\
    \bottomrule
  \end{tabular}
  \space
\end{small}
  \vspace{2mm}
\space
\end{table}

\FloatBarrier
\begin{figure}[!htbp]
  \centering
  \includegraphics[width=\textwidth]{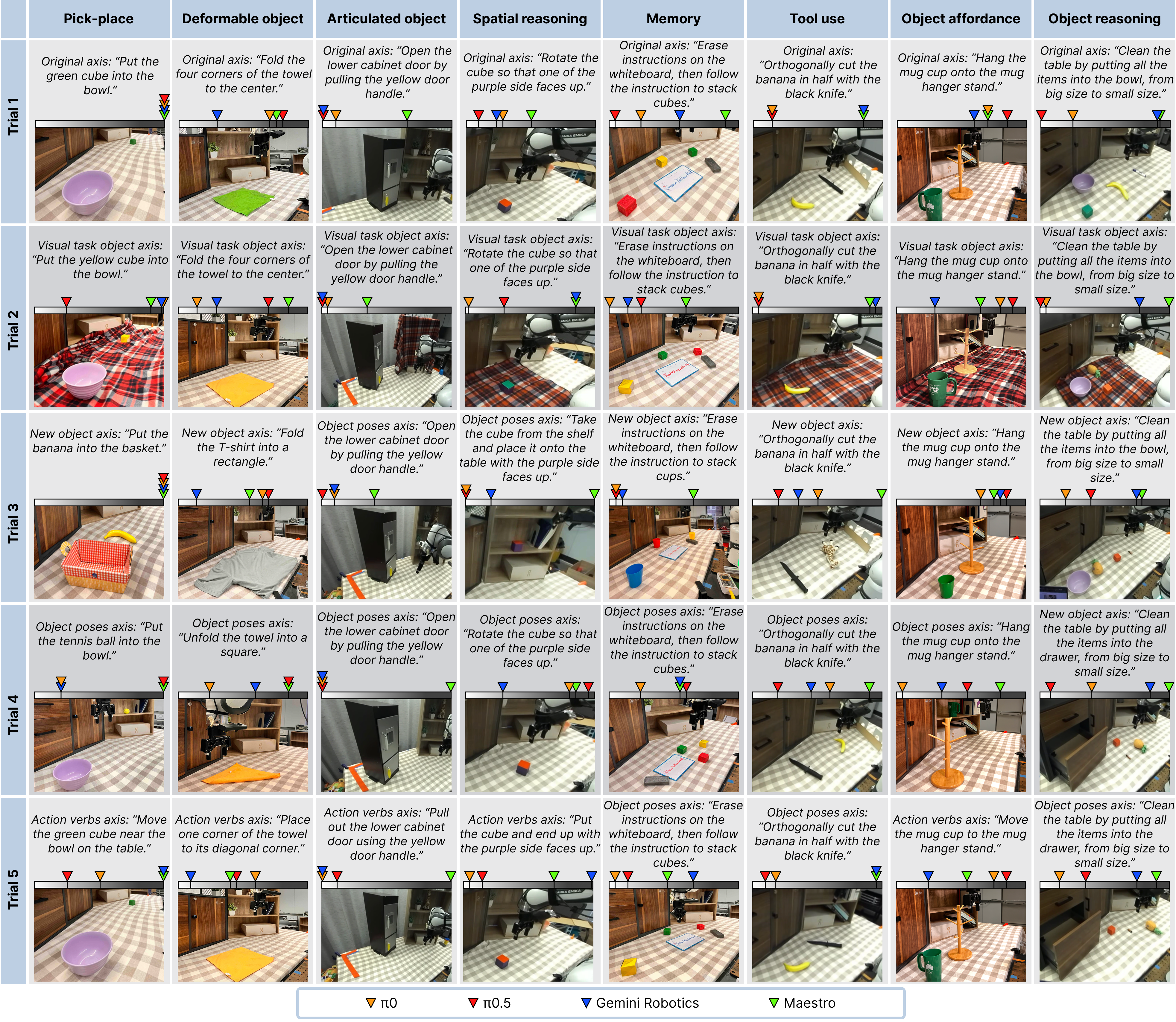}
  \caption{Per-trial visualization of tabletop manipulation results. Each row corresponds to a task and shows the original trial setting alongside the five STAR-Gen evaluation trials. Above each trial image, a gradient scale bar ranges from white (0 progress) on the left to black (100 progress) on the right. Colored triangular markers ($\blacktriangledown$) on this bar denote each method's score on that trial, with colors corresponding to the methods indicated in the legend.}
  \label{fig:stargen_trials}
\end{figure}

Fig.~\ref{fig:stargen_trials} visualizes every individual trial of our tabletop manipulation experiment. For each task, the figure shows the original trial setting alongside the semantic perturbations introduced by STAR-Gen~\cite{2025stargen}, which systematically varies object instances, object poses, scene backgrounds, and language instructions across trials. It also shows the resulting performance of each method on every trial. This per-trial visualization provides a more granular view than the aggregate statistics in Fig.~\ref{fig:main-table} and Table~\ref{tab:tabletop_results}, revealing how each method responds to specific types of variation.
\FloatBarrier

Across the evaluated settings, \methodname{} has the highest mean progress on six of eight tasks, while trial-level outcomes vary by task and setting. In contrast, $\pi_{0}$ and $\pi_{0.5}$ exhibit high variance across trials, with their markers scattered widely along the scale bar. In the \emph{Pick--Place} column (leftmost), the VLA baselines achieve perfect scores in some evaluated settings but substantially lower scores in others. This variation is reflected in their larger standard errors in Table~\ref{tab:tabletop_results}: $\pm 16.6$ and $\pm 18.4$ for $\pi_{0}$ and $\pi_{0.5}$, compared to $\pm 2.0$ for \methodname{}. On open cabinet, the baselines have low mean progress ($2.0$--$5.0$) and small SEs ($2.0$--$2.4$), reflecting uniformly low progress. On rotate cube, \methodname{} achieves $88.0 \pm 5.6$, compared with $52.0 \pm 16.5$ for the Gemini Robotics Agent, $23.0 \pm 15.0$ for $\pi_0$, and $30.0 \pm 17.0$ for $\pi_{0.5}$; its mean is higher and its SE is lower than those of the evaluated baselines. \space

\FloatBarrier
\section{Leave-One-Out Module Ablations}\label{app:loo_ablation}

This section reports the per-module leave-one-out ablations referenced in Sec.~\ref{sec:scaling_ablation}. Whereas the main-text scaling study \emph{adds} tool groups on top of Base CaP, here we start from the full \methodname{} and \emph{remove} one module category at a time, keeping all other components fixed.

\begin{table}[t]
  \centering
  \caption{Leave-one-out ablation on the Fold Towel and Rotate Cube tasks. Values are mean task progress out of 100 $\pm$ standard error (SE), with $n=5$ trials per method and task and $\mathrm{SE}=\mathrm{SD}/\sqrt{5}$. The full \methodname{} baseline reuses Table~\ref{tab:tabletop_results}; ablation SEs are calculated from the reported rounded SDs.}
  \label{tab:ablation_results}
  \setlength{\tabcolsep}{3pt}
  \renewcommand{\arraystretch}{1.2}
  \rowcolors{2}{gray!15}{white} %
  \begin{small}
  \begin{tabular}{>{\raggedright\arraybackslash}p{3cm}
                  >{\centering\arraybackslash}p{2cm}
                  >{\centering\arraybackslash}p{2cm}}
    \toprule
    \textbf{Method} & \textbf{Fold Towel} & \textbf{Rotate Cube} \\
    \midrule
    \methodname & $68.0 \pm 8.9$ & $88.0 \pm 5.6$ \\
    \methodname w/o advanced perception & $40.0 \pm 3.2$ & $25.0 \pm 0.0$ \\
    \methodname w/o geometry modules & $67.5 \pm 1.6$ & $42.5 \pm 14.2$ \\
    \bottomrule
  \end{tabular}
  \end{small}
\end{table}

While Table~\ref{tab:comparison} details the comprehensive list of improvements and additions we made over Base CaP, here we focus on two key categories of design choices among them. \textbf{\methodname w/o advanced perception} evaluates the impact of the \textit{task-relevant keypoint} and \textit{active perception} modules; \textbf{\methodname w/o geometry modules} tests the impact of the \textit{geometry} and \textit{linear algebra} modules. For systematic comparison, we fairly isolate each factor by following a ``change one at a time'' approach, creating variants where only one module category is reverted to the Base-CaP baseline while keeping all other components identical to \methodname.

\noindent \textbf{Task.} We use the ``\textbf{fold towel}'' and ``\textbf{rotate cube}'' tasks. In fold towel, the robot must fold the corners into the center, so the task requires reasoning about object geometry and specific points of interaction on the object. In rotate cube, the robot must rotate the cube until a particular color side faces up, requiring extensive spatial reasoning about rotation and object affordance.

\textbf{Results.} Table~\ref{tab:ablation_results} indicates task-dependent effects of the removed module categories. Removing advanced perception reduces the mean score on fold towel, which requires identifying and interacting with towel corners. The geometry removal has little effect on fold towel in these trials, but reduces mean progress on rotate cube. Both rotate-cube ablations still achieve partial progress. These results support the usefulness of the modules for the evaluated tasks.

\FloatBarrier
\section{Tool-Use Trace Analysis}\label{app:tooluse}

\begin{figure}[t]
  \centering
  \includegraphics[width=\linewidth]{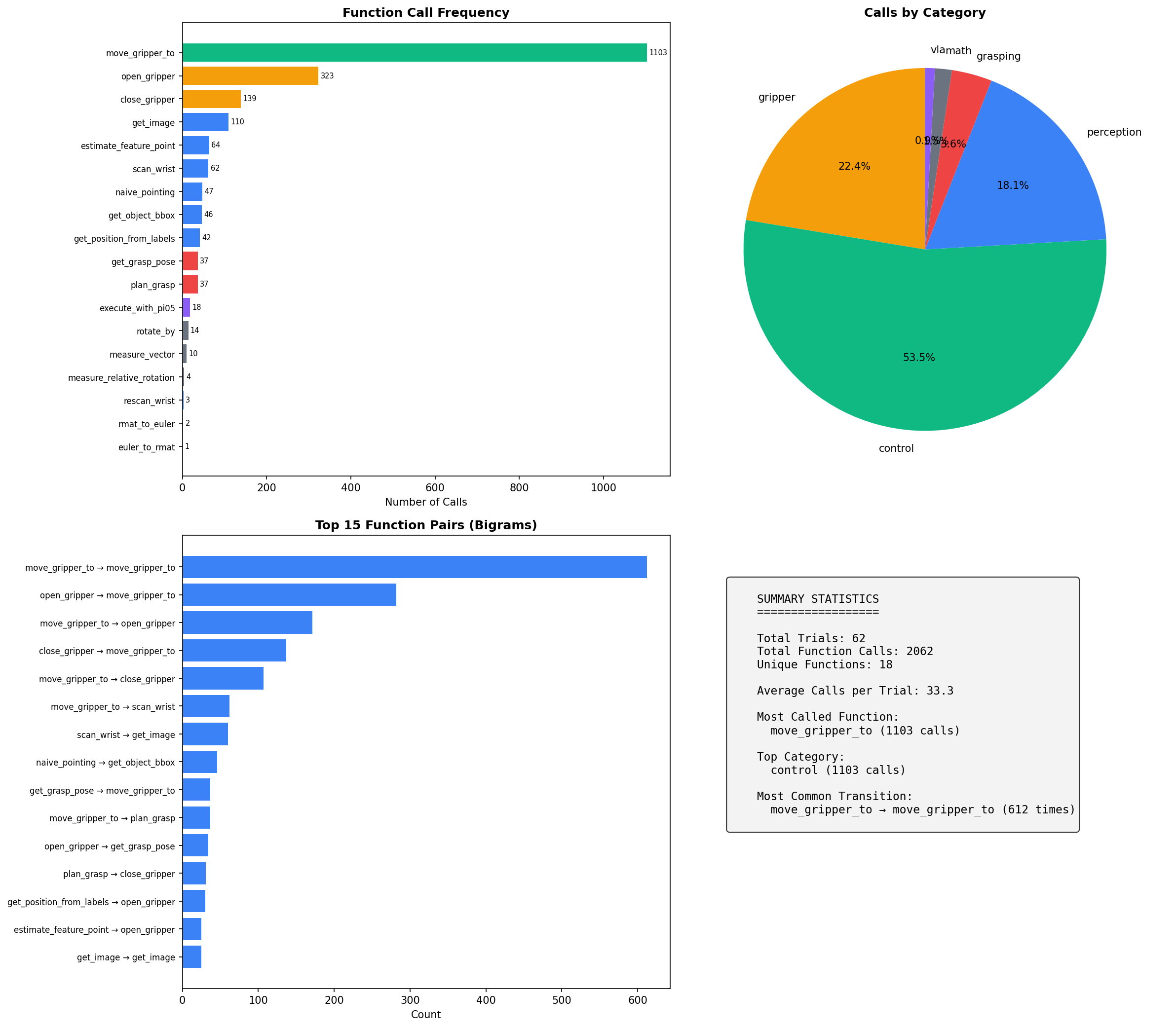}
  \caption{\textbf{Aggregate tool-use statistics across 62 \methodname{} tabletop trials} (2{,}062 calls across 18 modules in 6 categories). Top left: per-module call frequencies. Top right: calls grouped by module category.
Bottom left: most frequent consecutive tool-call pairs.
 \space
The full interactive passmap with per-task filtering is hosted on our \href{https://maestro-robot.github.io/}{project site}.}
  \label{fig:tooluse_summary}
\end{figure}

\begin{figure*}[t]
  \centering
  \includegraphics[width=0.95\textwidth]{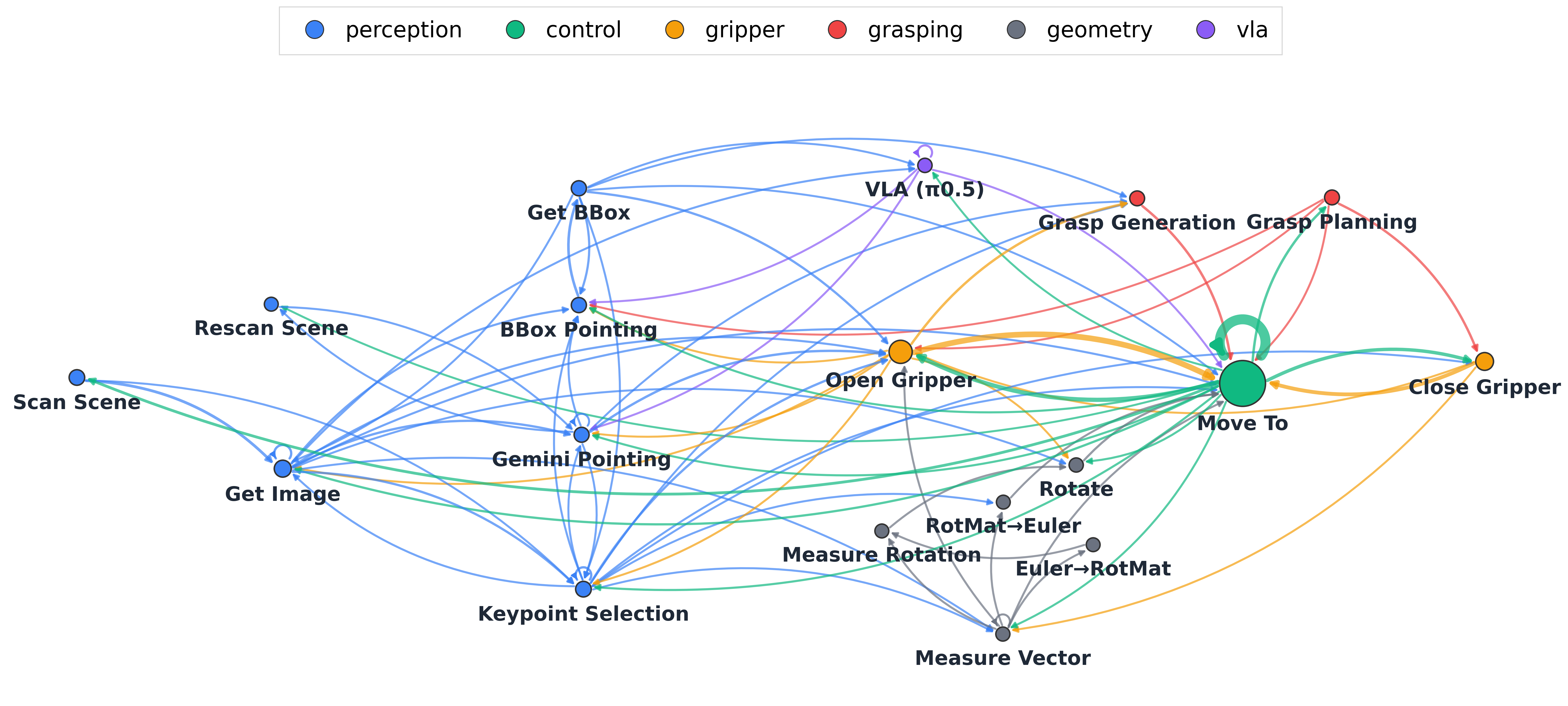}
  \caption{\textbf{Full \methodname{} tool-use passmap} aggregated across 62 tabletop trials (2{,}062 calls, 18 modules). Nodes are tool modules color-coded by category (perception, control, gripper, grasping, geometry, VLA); node size is proportional to call count and edge width to transition frequency. The interactive version on our \href{https://maestro-robot.github.io/}{project site} supports per-task filtering, drag-and-drop layouts, and highlighting of common workflows.}
  \label{fig:passmap_full}
\end{figure*}

\begin{figure}[t]
  \centering
  \includegraphics[width=\linewidth]{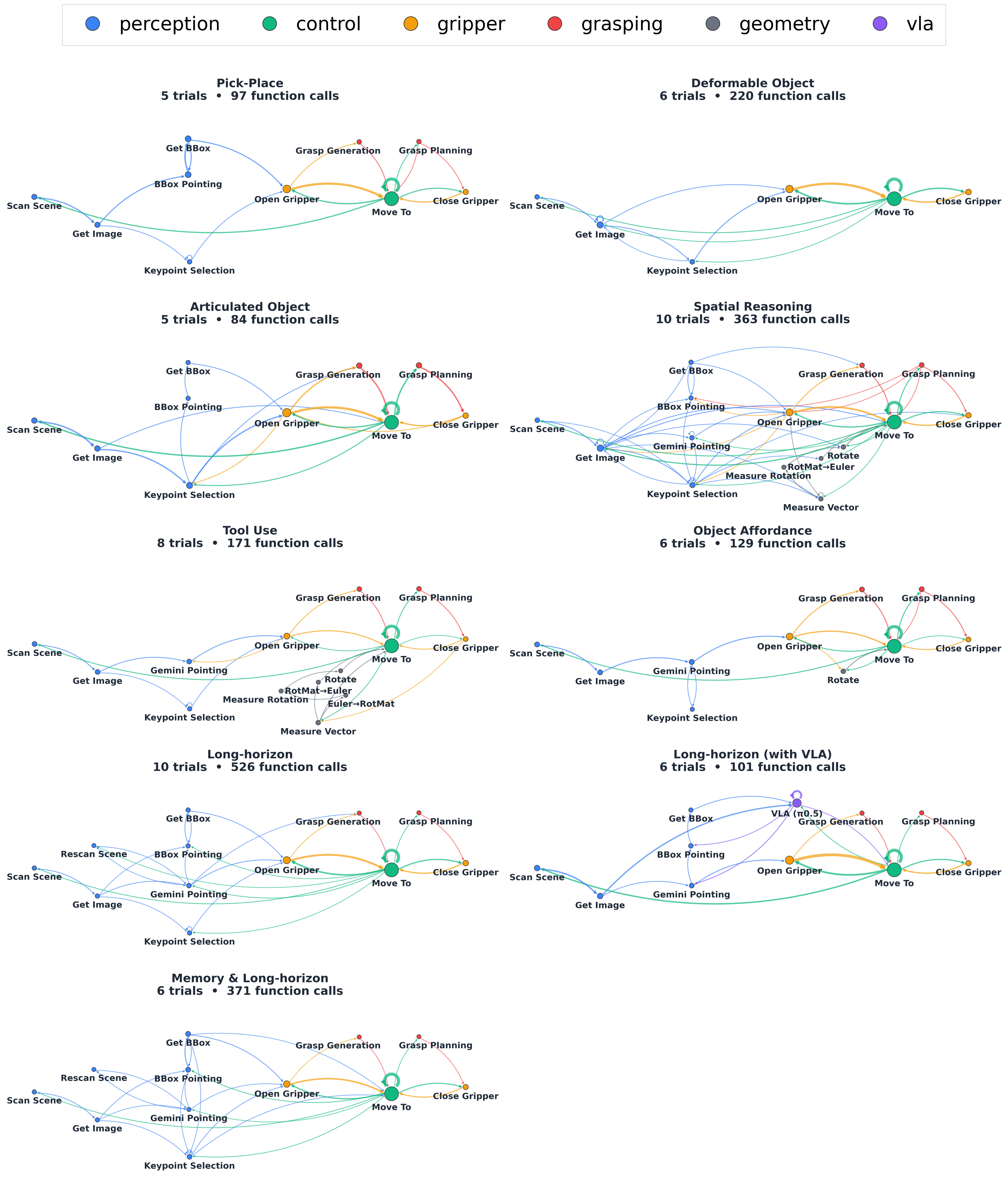}
  \caption{\textbf{Per-task tool-use passmap thumbnails.} Each panel shows the call graph for one task category (e.g.\ pick-place, deformable, articulated, long-horizon-with-VLA); node size encodes call count and edge width encodes transition frequency. The dominant edges and most-called modules differ across categories, illustrating that \methodname{} composes task-specific workflows rather than selecting a fixed pipeline. Full-resolution per-task passmaps and an interactive viewer with task-level filtering are available on our \href{https://maestro-robot.github.io/}{project site}.}
  \label{fig:tooluse_per_task}
\end{figure}

To verify that \methodname{} does more than execute a single canonical pipeline, we logged \methodname's tool calls during its tabletop trials. We recorded 2{,}062 tool calls drawn from 18 modules across 62 trials spanning 9 task categories, summarized in Fig.~\ref{fig:tooluse_summary} and
rendered as a graph of tool calls and their transitions (a ``passmap'') in Fig.~\ref{fig:passmap_full}
. The per-task call graphs (Fig.~\ref{fig:tooluse_per_task}) show that dominant tool sequences differ across task categories and are revised after failures. This pooled analysis includes the VLA-as-tool study in App.~\ref{app:vla_tool}. An interactive version of the passmap, with task-level filtering, free-form node layout, and saved views, is hosted on our \href{https://maestro-robot.github.io/}{project site}.

\noindent \textbf{Workflows are task-conditioned, not universal.} The most-called modules and the dominant transitions differ across categories: deformable tasks rely on \texttt{estimate\_feature\_point} cycles for keypoint re-tracking; articulated tasks chain \texttt{naive\_pointing} $\rightarrow$ \texttt{get\_object\_bbox} $\rightarrow$ \texttt{plan\_grasp} $\rightarrow$ \texttt{move\_gripper\_to}; long-horizon tasks add \texttt{rescan\_wrist} $\rightarrow$ \texttt{get\_image} re-perception cycles between subtasks. Fig.~\ref{fig:tooluse_per_task} makes this contrast visible at the graph level: it is the graph topology, not just the call counts, that changes with the task, so \methodname{} composes a task-specific workflow rather than selecting a fixed pipeline.

\noindent \textbf{Replanning produces observable in-trial corrections.}
Separately from the 62-trial tool-call graphs above, we analyze logged subtask attempts across 71 trials to quantify within-trial corrections. Of 63 initially failed subtask instances, 47 later received a logged success judgment (74.6\%).
 \space
This conditional, log-derived statistic is distinct from human-scored trial completion; Table~\ref{tab:subtask_corrections} reports the per-task breakdown. The recovered ``erase and stack'' trial, in which \methodname{} detects an incomplete erase and re-enters the wiping subroutine before continuing, is traced at the tool-call level below.

\begin{table}[H]
  \centering
  \caption{Per-task rates of later logged success after an initial failure. For each task category, we report the number of trials, the number of subtask instances whose first attempt failed ($X$), the subset with a later logged success judgment ($Y$), and the conditional rate $Y/X$. A ``subtask'' is a unique step index within a trial; inclusion in $Y$ requires a later attempt of the same step index in the same trial to reach \texttt{Success: True}. These log-derived judgments are distinct from human-scored trial completion.}
  \label{tab:subtask_corrections}
  \setlength{\tabcolsep}{6pt}
  \renewcommand{\arraystretch}{1.2}
  \resizebox{\textwidth}{!}{%
  \begin{tabular}{lcccc}
    \toprule
    \textbf{Task category} & \textbf{Trials} & \textbf{First-attempt failures} ($X$) & \textbf{Later logged success} ($Y$) & \textbf{Rate} ($Y/X$) \\
    \midrule
    Pick-place              &  7 &  2 &  1 & 50.0\% \\
    Deformable              &  6 &  3 &  3 & 100.0\% \\
    Articulated             &  5 &  2 &  2 & 100.0\% \\
    Tool use                &  7 &  8 &  4 & 50.0\% \\
    Spatial reasoning       &  9 & 14 & 12 & 85.7\% \\
    Object affordance       &  5 &  4 &  2 & 50.0\% \\
    Long-horizon            & 10 & 15 & 11 & 73.3\% \\
    Long-horizon (with VLA) &  6 &  7 &  6 & 85.7\% \\
    Memory \& long-horizon  & 16 &  8 &  6 & 75.0\% \\
    \midrule
    \textbf{Total}          & \textbf{71} & \textbf{63} & \textbf{47} & \textbf{74.6\%} \\
    \bottomrule
  \end{tabular}
  }%
\end{table}

\noindent \textbf{Trace case study: ``erase instructions on the whiteboard, then stack cubes'' with recovery.} \methodname{}'s opening tool sequence is \texttt{scan\_scene} $\rightarrow$ \texttt{get\_image} $\rightarrow$ \texttt{naive\_pointing} $\rightarrow$ \texttt{plan\_grasp} $\rightarrow$ \texttt{move\_gripper\_to} for the eraser. After the first wipe pass, the post-execution reflection (cued by the wrist-camera video) flags residual ink, and \methodname{} re-enters the wiping subroutine (\texttt{rescan\_wrist} $\rightarrow$ \texttt{get\_image} $\rightarrow$ \texttt{naive\_pointing} $\rightarrow$ \texttt{move\_gripper\_to}) before transitioning to the stacking subtask. The corresponding rollout, including the reflection prompt and the revised plan, is available in the Agent Dashboard on our \href{https://maestro-robot.github.io/}{project site} (``Erase (with replan)'').

\FloatBarrier
\section{Detailed Failure-Mode Analysis}\label{app:failure_cases}

\methodname{}'s failures concentrate in a few module classes rather than in high-level planning: imprecise pointing and segmentation under difficult lighting, motion-planning infeasibility and over-conservative collision checking, geometrically valid but task-inappropriate grasps, and, on the mobile platform, occasional default navigational fallbacks. This appendix enumerates the full per-embodiment failure modes, with concrete trial examples, below.

The Agent Dashboard on our \href{https://maestro-robot.github.io/}{project site} provides interactive trial replays that show how some of these failure modes surface in \methodname's plan-react-replan loop, including a recovered case where \methodname detects an incomplete erase and replans autonomously.

\par
\Needspace{5\baselineskip}
\noindent \textbf{Mobile platform}
\begin{enumerate}
    \space
\item \textbf{False Negatives.} In \textit{collect all toys}, YOLO never detected the strawberry plushy, so the system reported the task 100\% complete.
    \item \textbf{Color Confusion.} In \textit{collect all toys}, the system misidentified a red plushy as pink.
    \item \textbf{Lighting and Motion Sensitivity.} In \textit{press button}, motion blur combined with sunlight produced a ``bad image'' for the pointing module, sending the robot to the wrong position when approaching the door button.
    \item \textbf{Pointing Precision.} In \textit{throw garbage}, the VLM did not point to a reasonable location on the garbage can, so the dropped object landed half-on the rim.
    \item \textbf{Global-to-Local Viewpoint Selection.} The agent struggles to switch from macro-navigation (using the global 3D semantic map) to micro-manipulation. In \textit{throw garbage}, the initial observation correctly noted that the garbage can was on the floor, but on arrival the VLM defaulted to a ``FRONT'' camera pose and never pitched ``DOWN,'' so the short can was invisible at execution time.
    \item \textbf{Invalid Pose Generation.} The kinematic solver occasionally fails to produce a feasible end-effector orientation. In \textit{throw garbage}, no valid release pose was found at the end of the trajectory, preventing the robot from dropping the green plush ball into the can.
    \item \textbf{Default Navigational Fallback.} When given new action verbs, the VLM falls back on default navigational behaviors instead of adapting to the verb's meaning. In Task 3 Trial 5, the instruction was modified to ``hand it to the nearest person''; rather than visually searching for a human, the VLM returned to map origin $[0, 0, 0]$.
\end{enumerate}

\noindent \textbf{Tabletop platform}
\begin{enumerate}
    \item \textbf{Incomplete Collision Modeling of Grasped Objects.} cuRobo~\cite{curobo_report23} models the arm but not the grasped object, so the held object can still collide with surroundings in clutter even though the arm itself stays clear.
    \item \textbf{Pointing Inaccuracy.} Both VLM-based pointing (Gemini Robotics-ER 1.5) and segmentation-based pointing (Grounded-SAM2) lack the precision to localize fine-grained regions reliably, especially when multiple points are needed. VLM-based pointing also confuses certain color pairs (red/yellow, blue/yellow); segmentation-based pointing degrades under bright lighting.

\noindent When a perception error is visible in the returned observations, \methodname{} can acquire a new view, repeat perception, and replan. This does not cover undetected errors, which can propagate into later actions. The tool interface allows improved perception modules to replace existing ones without retraining the orchestration policy.
    \space
\item \textbf{Motion Planning Infeasibility.} VLM-selected end-effector poses are sometimes kinematically unreachable or unsolvable by cuRobo. cuRobo's collision checker is also over-conservative, flagging intended contacts (e.g., cutting, insertion) and blocking the interaction.
    \item \textbf{Insufficient Semantic Grasp Selection.} GraspGen~\cite{murali2025graspgen} returns geometrically valid grasps but does not reason about task affordances, so the robot may grasp a knife by the blade or a mug by an orientation that prevents the planned hang.
    \item \textbf{Limited Spatial Reasoning for Complex Affordances.} Tasks requiring precise relative-orientation reasoning, such as hanging a mug on a holder, remain hard: the VLM fails to align the handle opening with the holder branch and to leave clearance along the insertion direction, often knocking the holder over. The available geometry and vector tools are not consistently leveraged for these multi-step alignments.
    \item \hspace{0pt}\textbf{VLM Inference Latency.} Each plan-react-replan iteration calls the VLM, making \methodname considerably slower than end-to-end VLA policies such as $\pi_{0.5}$. Reducing VLM inference latency and overall compute cost remains an important direction for future work.

\end{enumerate}

\subsection{Latency breakdown}\label{app:latency}
\begin{table}[H]
  \centering
  \small
  \setlength{\tabcolsep}{3pt}
  \caption{Component latency aggregated across a mixture of main and VLA experiments. The listed tool calls account for approximately 47\% of total experiment wall time; the breakdown is not exhaustive, with remaining time including robot execution, waiting, logging, and other operations. Counts denote tool invocations, not independent trials; initial planning and replanning subdivide VLM planning and are not counted twice.}
  \label{tab:latency}
  \begin{tabular}{lrrrrrr}
    \toprule
    Component & Total (min) & Wall (\%) & Calls & Mean (s) & Min (s) & Max (s) \\
    \midrule
    VLM planning & 152.8 & 22.1 & 570 & 16.1 & 5.1 & 77.6 \\
    \quad Initial planning & 21.1 & 3.1 & 61 & 20.8 & 9.6 & 40.3 \\
    \quad Replanning & 131.7 & 19.1 & 509 & 15.5 & 5.1 & 77.6 \\
    GroundedSAM segmentation & 104.6 & 15.1 & 1433 & 4.4 & 0.9 & 24.8 \\
    Keypoint picking & 37.9 & 5.5 & 59 & 38.5 & 8.6 & 124.8 \\
    Gemini pointing & 14.3 & 2.1 & 41 & 20.9 & 3.7 & 39.4 \\
    $\pi_{0.5}$ policy rollout & 13.7 & 2.0 & 17 & 48.4 & 22.1 & 69.8 \\
    Qwen success check & 0.6 & 0.09 & 151 & 0.25 & 0.24 & 0.32 \\
    Motion planning & 0.4 & 0.1 & 37 & 0.6 & 0.0 & 0.8 \\
    Grasp generation & 0.3 & 0.0 & 74 & 0.3 & 0.0 & 0.7 \\
    \bottomrule
  \end{tabular}
\end{table}

\section{\methodname{} with a VLA Tool}\label{app:vla_tool}

This appendix gives the full setup, baselines, and per-method results for the VLA-as-tool experiment summarized in App.~\ref{sec:vla_tool}, isolating the contribution of $\pi_{0.5}$ as a callable pick-and-place module within \methodname's toolset.

\noindent \textbf{Setup.} Here we evaluate $\pi_{0.5}$ as a callable tool within \methodname{}. We do not invoke this tool in the zero-shot results or tool/harness ablations (Secs.~\ref{sec:tabletop_results}--\ref{sec:scaling_ablation}). To isolate the contribution of this module, we evaluate on \textbf{clean up the table, largest object first}, a long-horizon task that combines repeated pick-and-place steps with size-based semantic ordering. We compare \methodname with its full toolset (\textbf{\methodname w/ $\pi_{0.5}$}) against an ablation that does not have access to $\pi_{0.5}$ (\textbf{\methodname w/o $\pi_{0.5}$}) and must execute every sub-skill through its own primitive controllers. Table~\ref{tab:maestro_vla_results} reports average task progress and average wall-clock completion time.

\begin{table}[H]
  \centering
  \caption{Evaluation of \methodname{} with and without $\pi_{0.5}$ as a callable pick-and-place tool, on the \emph{clean up the table, largest object first} task. Reported are average task progress (0--100; higher is better) with $\pm1$ standard error across five trials per method ($\mathrm{SE}=\mathrm{SD}/\sqrt{5}$, in percentage points), and average wall-clock completion time.}
  \label{tab:maestro_vla_results}
  \space
\setlength{\tabcolsep}{4pt}
  \renewcommand{\arraystretch}{1.2}
  \rowcolors{2}{gray!15}{white}
  \begin{small}
  \begin{tabular}{>{\raggedright\arraybackslash}p{2.6cm}cc}
    \toprule
    \textbf{Method} & \textbf{Progress (\%)} & \textbf{Time} \\
    \midrule
    \methodname w/o $\pi_{0.5}$ & $89 \pm 5.1$ & $11'\,18''$ \\
    \methodname w/ $\pi_{0.5}$ & $94 \pm 4.9$ & $4'\,59''$ \\
    $\pi_{0.5}$ & $17 \pm 9.1$ & $2'\,16''$ \\
    $\pi_{0}$ & $21 \pm 6.6$ & $2'\,58''$ \\
    Gemini Robotics & $80 \pm 3.0$ & $11'\,53''$ \\
    \bottomrule
  \end{tabular}
  \end{small}
\end{table}

\noindent \textbf{Results.} $\pi_{0}$ and $\pi_{0.5}$ alone reach only 21\% and 17\% progress, as neither can sustain the size-based ordering across the task. \methodname w/o $\pi_{0.5}$ already reaches 89\%, and \methodname w/ $\pi_{0.5}$ further improves to 94\% while cutting completion time from $11'\,18''$ to $4'\,59''$ (a 2.3$\times$ speedup). The gain comes from \methodname offloading short pick-and-place steps to $\pi_{0.5}$'s feedforward inference while retaining its own long-horizon planning and reactive replanning. VLAs and \methodname are thus complementary: VLAs serve as fast, reliable skill modules, and \methodname provides the orchestration to deploy them on tasks they cannot solve in isolation. Rollout videos for this experiment are available on our \href{https://maestro-robot.github.io/}{project site}.

\section{Automatic Data Generation: Tether Case Study}\label{app:tether}

This section provides additional background on the Tether framework~\cite{liang2026tether} used in our autonomous data collection experiments (App.~\ref{sec:tether_exp}).

Tether is a framework for automatic robot data generation: starting from as few as 10 recorded execution rollouts, it adapts their trajectories to new scene configurations to generate a larger, diverse dataset. The key insight is \emph{correspondence-driven trajectory warping}: given a recorded rollout in one scene configuration, Tether extracts a small set of task-critical waypoints from that rollout (end-effector poses at frames where the gripper open/close state toggles) and uses DINOv2/Stable-Diffusion-based semantic correspondences to locate the matching waypoints in a new scene (e.g., with different object poses or instances). The 2D matches are backprojected to 3D via stereo, yielding per-waypoint displacements that are linearly interpolated along the recorded trajectory to warp the full end-effector path (timing rescaled to preserve speed), producing a new trajectory that accomplishes the same task under different conditions. We refer the reader to \cite{liang2026tether} for full details on the correspondence computation and warping procedure.

In our experiments, \methodname{} supplies the initial execution rollouts in place of human teleoperation. Tether then uses these recordings to generate data across new scene configurations. Only environment resets are manual: between rollouts, a human resets the workspace with deliberately varied object placements (and, where applicable, object instances), and Tether adapts a recorded trajectory to each new configuration, so the dataset grows by one trajectory per reset.

We detail below the autonomous data-collection experiment summarized in App.~\ref{sec:tether_exp}.

\begin{figure}[H]
  \centering
  \includegraphics[width=0.55\linewidth]{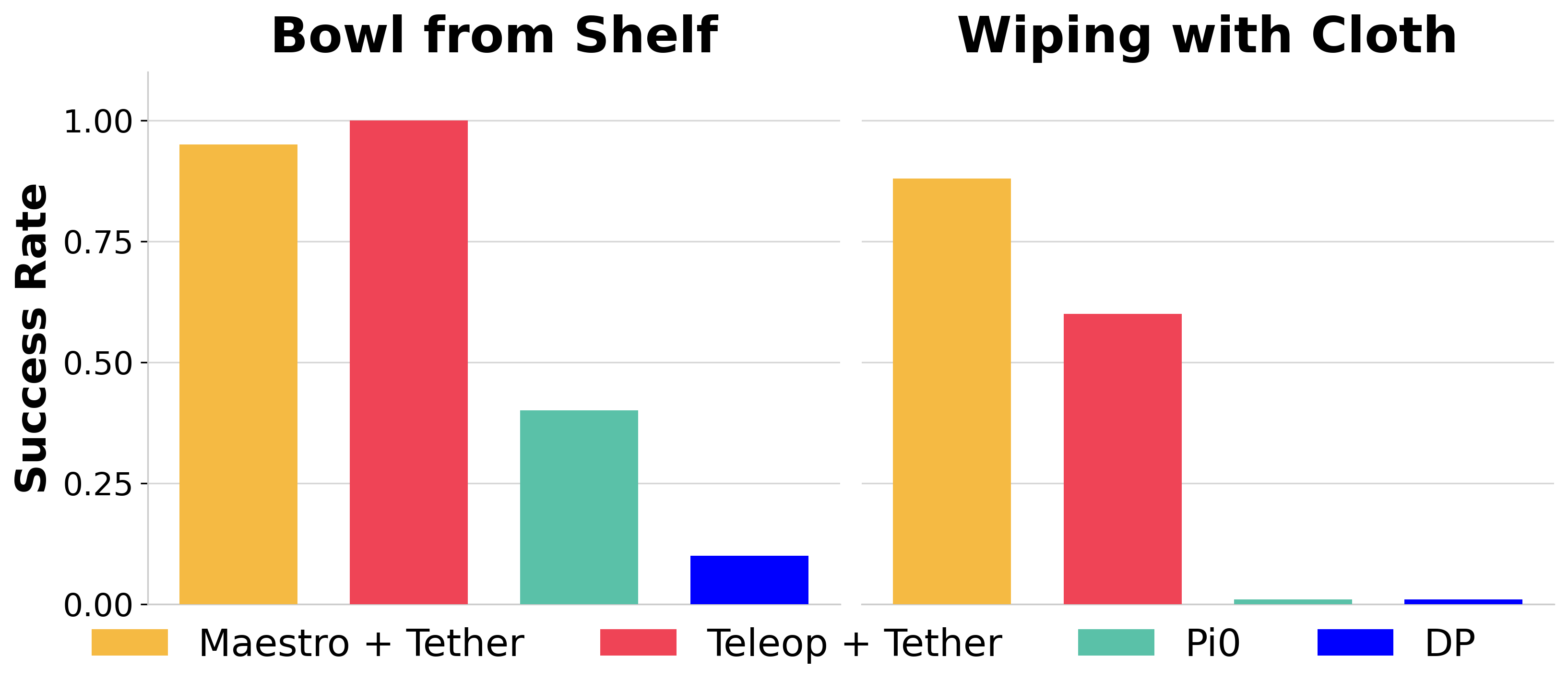}

  \caption{
\textbf{Automatic data generation with Tether~\cite{liang2026tether}: a two-task case study.} We compare \emph{Bowl from Shelf} and \emph{Wiping with Cloth}. \methodname{} + Tether starts from 10 \methodname{} execution rollouts; Tether starts from 10 human-teleoperated rollouts, and Diffusion Policy~\cite{chi2023diffusionpolicy} trains on Tether's human-teleoperated data.
\space
The $\pi_0$~\cite{black2024pi_0} baseline is evaluated zero-shot. See the \href{https://maestro-robot.github.io/}{project website} for videos of the data-collection tasks.
}
  \space
\label{fig:data_collection}
\space
\end{figure}

\textbf{Setup.} We evaluate on two tasks from the Tether benchmark: \emph{Bowl from Shelf} (retrieving a bowl from a shelf) and \emph{Wiping with Cloth} (wiping a surface with a cloth). We compare four methods: (1)~\textbf{\methodname + Tether}, where Tether starts from 10 \methodname{} execution rollouts and generates data across new scene configurations; (2)~\textbf{Tether}, which instead starts from 10 human-teleoperated rollouts; (3)~\textbf{$\pi_0$}~\cite{black2024pi_0}, a state-of-the-art VLA run zero-shot; and (4)~\textbf{Diffusion Policy (DP)}~\cite{chi2023diffusionpolicy}, trained on Tether's human-teleoped demonstrations.

\textbf{Results.} As shown in Fig.~\ref{fig:data_collection}, on these two tasks \methodname + Tether achieves performance comparable to human-teleoped Tether (e.g., 95\% vs.\ 100\% on Bowl from Shelf, and 88\% on Wiping with Cloth),
and both have higher reported success rates than $\pi_0$ and Diffusion Policy.
 \space
These results provide preliminary evidence that \methodname{} execution rollouts can replace human-teleoperated recordings as the starting point for automatic data generation with Tether. This pipeline requires no human teleoperation, although environment resets remain manual. More broadly, this suggests that \methodname{} could serve as an autonomous data source for downstream policy learning. See our \href{https://maestro-robot.github.io/}{project site} for videos of the Tether data collection experiments.

\section{Monitor Reliability Audit}\label{app:monitor_audit}
We audited the main VLM replanning judgments against human review of the corresponding trial videos. The audit contains 61 trials with usable logs and 509 logged judgments. We excluded 137 post-completion idle judgments, yielding 372 evaluated judgments; the exclusion rule omits steps with no robot command and includes 26 reviewer overrides.

Treating a claim of step success as positive, the counts are 204 true positives, 36 false positives, 130 true negatives, and 2 false negatives. Accuracy is $334/372=89.8\%$, success precision is $204/240=85.0\%$, and success recall is $204/206=99.0\%$. Among positive judgments asserting task completion, $22/55=40.0\%$ are incorrect; among positive intermediate-step judgments, $14/185=7.6\%$ are incorrect.

\end{document}